\documentclass[12pt]{article}
\usepackage[margin=1in]{geometry}
\usepackage{setspace}
\usepackage{enumitem}
\usepackage{amssymb}
\usepackage[round]{natbib}
\usepackage{amsmath,amsfonts,bm}

\usepackage{amsthm}
\usepackage{microtype}
\usepackage{graphicx}
\usepackage{booktabs} 
\usepackage[toc,page]{appendix}
\usepackage{comment}
\usepackage{caption}
\usepackage{subcaption}
\usepackage[dvipsnames]{xcolor}

\newtheorem{theorem}{Theorem}

\newcommand{\E}{\mathbb{E}}

\usepackage{hyperref}

\usepackage{algorithm}
\usepackage{algpseudocode}

\usepackage{authblk}

\usepackage[utf8]{inputenc}

\begin{document}

\title{\bf Inferential Wasserstein Generative Adversarial Networks}
\author{Yao Chen, Qingyi Gao, Xiao Wang}
\affil{Department of Statistics, Purdue University}
\date{}

\maketitle


\begin{abstract}
Generative Adversarial Networks (GANs) have been impactful on many problems and applications but suffer from unstable training. The Wasserstein GAN (WGAN) leverages the Wasserstein distance to avoid the caveats in the minmax two-player training of GANs but has other defects such as mode collapse and lack of metric to detect the convergence. We introduce a novel inferential Wasserstein GAN (iWGAN) model, which is a principled framework to fuse auto-encoders and WGANs. The iWGAN model jointly learns an encoder network and a generator network motivated by the iterative primal dual optimization process. The encoder network maps the observed samples to the latent space and the generator network maps the samples from the latent space to the data space. We establish the generalization error bound of the iWGAN to theoretically justify its performance. We further provide a rigorous probabilistic interpretation of our model under the framework of maximum likelihood estimation. The iWGAN, with a clear stopping criteria, has many advantages over other autoencoder GANs. The empirical experiments show that the iWGAN greatly mitigates the symptom of mode collapse, speeds up the convergence, and is able to provide a measurement of quality check for each individual sample. We illustrate the ability of the iWGAN by obtaining competitive and stable performances for benchmark datasets.\medskip{}

\end{abstract}

\newpage

\section{Introduction}
One of the goals of generative modeling is to match the model distribution $P_\theta(x)$ with parameters $\theta$ to the true data distribution $P_X$ for a random variable $X\in {\cal X}$. For latent variable models, the data point $X$ is generated from a latent variable $Z\in {\cal Z}$ through a conditional distribution $p(x|z)$. Here ${\cal X}$ denotes the support for $P_X$ and ${\cal Z}$ denotes the support for $P_Z$. In this paper, we consider models with $Z\sim {\cal N}(0, I)$. There has been a surge of research on deep generative networks in recent years and the literature is too vast to summarize here \citep{kingma2013auto, goodfellow2014generative,li2015generative, gao20, qiu20jasa}. These models have provided a powerful framework for modeling complex high dimensional datasets. 

We start introducing two main approaches for generative modeling. The first one is called {\it variational auto-encoders} (VAEs) \citep{kingma2013auto}, which use variational inference \citep{blei17} to learn a model by maximizing the lower bound of the likelihood function. Specifically, let the latent variable $Z$ be drawn from  a prior $p(z)$ and the data $X$ have a likelihood $p(x|z)$ that is conditioned on $Z=z$. Unfortunately obtaining the marginal distribution of $X$ requires computing an intractable integral $p(x) = \int p(x|z)p(z)dz$. Variational inference approximates the posterior $p(z|x)$ by a family of distribution $q_\eta(z|x)$ with the parameter $\eta$. The objective is to maximize the lower bound of the log-likelihood function known as the evidence lower bound (ELBO). The ELBO is given by
\[
\mbox{ELBO} = \mathbb E_{z\sim q_\eta(z|x)} \big[\log p(x|z)\big] - \mathbb E_{z\sim q_\eta(z|x)}\big[\log {q_\eta(z|x)\over p(z)}\big].
\]
Note that the difference between the log-likelihood $\log p(x)$ and the ELBO is the Kullback-Leibler divergence between $q_\eta(z|x)$ and the true posterior $p(z|x)$. Usually $q_\eta(z|x)$ is a normal density where both the conditional mean and the conditional covariance are modeled by deep neural networks (DNNs), so that the first term of the ELBO can be approximated efficiently by Monte Carlo methods and the second term can be calculated explicitly. Therefore, the ELBO allows us to do approximate posterior inference with tractable computation. VAEs have elegant theoretical foundations but the drawback is that they tend to produce blurry images. The second approach is called {\it generative adversarial networks} (GANs) \citep{goodfellow2014generative}, which learn a model by using a powerful discriminator to distinguish between real data samples and generative data samples. Specifically, we define a generator $G: {\cal Z}\rightarrow {\cal X}$ and a discriminator $D: {\cal X}\rightarrow \mathbb [0, 1]$. The generator and discriminator play a two-player minimax game by being alternatively updated, such that the generator tries to produce real-looking images and the discriminator tries to distinguish between generated images and observed images. The GAN objective can be written as $\min_G\max_D V(G, D)$, where
\[
V(G, D) = \mathbb E_{x\sim P_X} \log D(x) + \mathbb E_{z\sim P_Z}\log(1 - D(G(z))).\]
GANs produce more visually realistic images but suffer from the unstable training and the mode collapse problem. Although there are many variants of generative models trying to take advantages of both VAEs and GANs \citep{tolstikhin2018wasserstein, rosca2017variational}, to the best of our knowledge, the model which provides a unifying framework combining the best of VAEs and GANs in a principled way is yet to be discovered.

\subsection{Related work}

In this section, we provide a brief introduction of different variants of generative models.

\noindent {\bf Wasserstein GAN.} The Wasserstein GAN (WGAN) \citep{arjovsky2017wasserstein} is an extension to the GAN that improves the stability of the training by introducing a new loss function motivated from the Waseerstein distance between two probability measures  \citep{villani2008optimal}. Let $P_{G(Z)}$ denote the generative model distribution through the generator $G$ and the latent variable $Z\sim P_Z$. Both the vanilla GAN \citep{goodfellow2014generative} and the WGAN can be viewed as minimizing certain divergence between the data distribution $P_X$ and the generative distribution $P_{G(Z)}$. For example, the Jensen-Shannon (JS) divergence is implicitly used in vanilla GANs \citep{goodfellow2014generative}, while the $1$-Wasserstein distance  is employed in WGANs. Empirical experiments suggest that the Wasserstein distance is a more sensible measure to differentiate probability measures supported in low-dimensional manifold. In terms of training, it turns out that it is hard or even impossible to compute these standard divergences in probability, especially when $P_X$ is unknown and $P_{G(Z)}$ is parameterized by DNNs. Instead, the training of WGANs is to study its dual problem because of the elegant form of Kantorovich-Rubinstein duality \citep{villani2008optimal}.

\noindent{\bf Autoencoder GANs.} 
The main difference between autoencoder GANs and standard GANs is that, besides the generator $G$, there is an encoder $Q: {\cal X}\rightarrow {\cal Z}$ which maps the data points into the latent space. This deterministic encoder is to approximate the conditional distribution $p(z|x)$ of the latent variable $Z$ given the data point $X$. \cite{larsen2016autoencoding} first introduced the VAE-GAN, which is a hybrid of VAEs and GANs and uses a GAN discriminator to replace a VAE’s decoder to learn the loss function. For both the Adversarially Learned Inference (ALI) \citep{dumoulin2017adversarially} and the Bidirectional Generative Adversarial Network (BiGAN) \citep{donahue2016adversarial}, the objective is to match two joint distributions, $(X, Q(X))$ and $(G(Z), Z)$, under the framework of vanilla GANs. When the algorithm achieves equilibrium, these two joint distributions roughly match. It is expected to obtain more meaningful latent codes by $Q(X)$, and this should improve the quality of the generator as well. For other VAE-GAN variants, please see \cite{rosca2017variational, mescheder2017adversarial, hu2017unifying, ulyanov2018takes}.

\noindent{\bf Energy-Based GANs.} Energy-based Generative Adversarial Networks (EBGANs) \citep{zhao2016energy} view the discriminator as an energy function that attributes low energies to the regions near the data manifold and higher energies to other regions, and the generator as being trained to produce contrastive samples with minimal energies.
\cite{han19} presented the joint training of generator model, energy-based model, and inference model, which introduces a new objective function called divergence triangle that makes the processes of sampling, inference, energy evaluation readily available without the need for costly Markov chain
Monte Carlo methods.

\noindent{\bf Duality in GANs.} 
Regarding the optimization perspectives of GANs, \citep{chen2018training, zhao2018information} studied duality-based methods for improving algorithm performance for training. 
\cite{farnia2018convex} developed a convex duality framework to address the case when the discriminator is constrained into a smaller class. \cite{grnarova2018evaluating} developed an evaluation metric to detect the non-convergence behavior of vanilla GANs, which is the duality gap defined as the difference between the primal and the dual objective functions. 

\subsection{Our Contributions}

Although there are many interesting  works on autoencoder GANs, it remains unclear what the principles are underlying the fusion of auto-encoders and GANs. For example, do there even exist these two mappings, the encoder $Q$ and the decoder $G$, for any high-dimensional random variable $X$, such that $Q(X)$ has the same distribution as $Z$ and $G(Z)$ has the same distribution as $X$? Is there any probabilistic interpretation such as the maximum likelihood principle on autoencoder GANs?  What is the generalization performance of autoencoder GANs? In this paper, we introduce inferential WGANs (iWGANs), which provide satisfying answers for these questions. We will mainly focus on the 1-Wasserstein distance, instead of the Kullback-Leibler divergence. We borrow the strength from both the primal and the dual problems and demonstrate the synergistic effect between these two optimizations. The encoder component turns out to be a natural consequence from our algorithm. The iWGAN learns both an encoder and a decoder simultaneously. We prove the existence of meaningful encoder and decoder, establish an equivalence between the WGAN and iWGAN, and develop the generalization error bound for the iWGAN. Furthermore, the iWGAN has a natural probabilistic interpretation under the maximum likelihood principle. Our learning algorithm is equivalent to the maximum likelihood estimation motivated from the variational approach when our model is defined as an energy-based model based on an autoencoder. As a byproduct, this interpretation allows us to perform the quality check at the individual sample level. In addition, we demonstrate the natural use of the duality gap as a measure of convergence for the iWGAN, and show its effectiveness for various numerical settings. Our experiments do not experience any mode collapse problem.

The rest of the paper is organized as follows. Section 2 presents the new iWGAN framework, and its extension to general inferential f-GANs. Section 3 establishes the generalization error bound and introduces the algorithm for the iWGAN. The probabilistic interpretation and the connection with the maximum likelihood estimation are introduced in Section 4. Extensive numerical experiments are demonstrated in Section 5 to show the advantages of the iWGAN framework. Proofs of theorems and additional numerical results are provided in the Appendix.

\section{The iWGAN Model}\label{sec:iWGAN}

The autoencoder generative model consists of two parts: an encoder $Q$ and a generator $G$. The encoder $Q$ maps a data sample $x\in {\cal X}$ to a latent variable $z\in {\cal Z}$, and the generator $G$ takes a latent variable $z\in {\cal Z}$ to produce a sample $G(z)$. In general, the autoencoder generative model should satisfy the following three conditions {\it simultaneously}: (a) The generator can generate images which have a similar distribution with observed images, i.e., the distribution of $G(Z)$ is similar to that of $P_X$; (b) The encoder can produce meaningful encodings in the latent space, i.e., $Q(X)$ has a similar distribution with $Z$; (c) The reconstruction errors of this model based on these meaningful encodings are small, i.e., the difference between $X$ and $G(Q(X))$ is small. 

We emphasize that the benefit of using an autoencoder is to encourage the model to better represent {\it all} the data it is trained with, so that it discourages mode collapse. We first show that, for any distribution residing on a compact smooth Riemannian manifold \footnote{A smooth manifold ${\cal X}$ is a manifold with a $C^{\infty}$ atlas on ${\cal X}$. A $C^{\infty}$ atlas is a collection of charts $\{\varphi_\alpha: U_\alpha \rightarrow \mathbb R^d\}$ such that $\{U_\alpha\}$ covers ${\cal X}$, and for all $\alpha$ and $\beta$, the transition map $\varphi_\alpha \cdot \varphi_\beta^{-1}$ is a $C^{\infty}$ map.  Here $U_\alpha$ is an open subset of ${\cal X}$. For any point $p\in {\cal X}$, let $T_p{\cal X}$ be the tangent space of ${\cal X}$ at $p$. A Riemannian metric assigns to each $p$ a positive definite inner product $g_p: T_p{\cal X} \times T_p{\cal X} \rightarrow \mathbb R$, along with which comes a norm $\|\cdot\|_p: T_p{\cal X} \rightarrow \mathbb R$ defined by $\|v\|_p = \sqrt{g_p(v, v)}$.  The smooth manifold ${\cal X}$ endowed with this metric $g$ is called a smooth Reimannian manifold.}, there always exists an encoder $Q^*: {\cal X}\rightarrow {\cal Z}$ which guarantees meaningful encodings and exists a generator $G^*: {\cal Z}\rightarrow {\cal X}$ which generates samples with the same distribution as data points by using these meaningful codes. 

\begin{theorem}\label{thm:basic}
Consider a continuous random variable $X\in {\cal X}$, where ${\cal X}$ is a $d$-dimensional compact smooth Riemannian manifold.  Then, there exist two mappings $Q^*: {\cal X}\rightarrow \mathbb R^{p}$ and $G^*: \mathbb R^{p} \rightarrow {\cal X}$, with $p=\max\{d(d+5)/2, d(d+3)/2+5\}$, such that $Q^*(X)$ follows a multivariate normal distribution with zero mean and identity covariance matrix and  $G^*\circ Q^*$ is an identity mapping, i.e., $X = G^*(Q^*(X))$.
\end{theorem}

Theorem \ref{thm:basic} is a natural consequence of the Nash embedding theorem \citep{nash56, gunther} and the probability integral transformation \citep{rosenblatt52}. In Theorem \ref{thm:basic}, we have proved the existence of $Q^*$ and $G^*$, however, learning $Q^*$ and $G^*$ from the data points is still a challenging task. Consider a general $f$-GAN model \citep{nowozin2016f}. Let $h:\mathbb R\rightarrow(-\infty, \infty]$ be a convex function with $h(1) = 0$. The $f$-GAN defines the $f$-divergence between the data distribution $P_X$ and the generative model distribution $P_{G(Z)}$ for the generator $G$ as:
\[
\text{GAN}_h(P_X, P_{G(Z)}) = \sup_{f\in {\cal F}}\Big[\mathbb E_{X}\left\{f(X)\right\} - \mathbb E_{Z}\left\{h^{*}(f(G(Z))\right\}\Big],
\]
where $h^{*}(x) = \sup_{y}\{x\cdot y-h(y)\}$ is the convex conjugate of $h$ and ${\cal F}=\{f|f:\cal X \rightarrow \mathbb R\}$ is a class of functions whose output range is the domain of $h^{*}$. When $f$ is approximated by a DNN, its output range can be achieved by choosing an appropriate activation function specific to the f-divergence used. For example, if $h(x)=x\log (x)-(x+1)\log (x+1)$, then the corresponding convex conjugate $h^{*}(x) = -\log(1-\exp(x))$. To satisfy the above condition, we select the output activation function of the DNN $f$ to be $\sigma(v) = -\log (1+\exp(-v))$ such that the $f$-GAN can recover the original vanilla GAN \citep{goodfellow2014generative}. If $h(x) = 0$ when $x = 1$ and $h(x) = \infty$ otherwise, we have $h^{*}(x) = x$. With the property that $\cal F$ is 1-Lipschitz function class, the $f$-GAN turns to be the WGAN.

For ease of presentation, we illustrate our methodology by mainly focusing on the Wasserstein distance and the inferential WGAN (iWGAN) model. The extension to general inferential f-GANs (ifGANs) is straightforward and will be presented in Section \ref{sec:fgan}.

\subsection{iWGAN}

Recall that the $1$-Wasserstein distance between $P_X$ and $P_{G(Z)}$ is defined as
\begin{equation}\label{eq:primal}
    W_1(P_X, P_{G(Z)}) = \inf_{\pi \in \Pi(P_X, P_Z)} \mathbb{E}_{(X, Z)\sim \pi} \big\|X- G(Z)\big\|,
\end{equation}
where $\|\cdot\|$ represents the $L_2$-norm and $\Pi(P_X, P_Z)$ is the set of all joint distributions of $(X, Z)$ with marginal measures $P_X$ and $P_Z$, respectively. The main difficulty  in (\ref{eq:primal}) is to find the optimal coupling $\pi$, and this is a constrained optimization because the joint distribution $\pi$ needs to match these two marginal distributions $P_X$ and $P_Z$.

Based on the Kantorovich-Rubinstein duality \citep{villani2008optimal}, the WGAN studies the $1$-Wasserstein distance (\ref{eq:primal}) through its dual format
\begin{equation}\label{eq:dual}
    W_1(P_X, P_{G(Z)}) = \sup_{f\in {\cal F}}\Big[ \mathbb E_{X\sim P_X} \big\{ f(X)\big\} - \mathbb E_{Z\sim P_Z} \big\{ f(G(Z))\big\}\Big],
\end{equation}
where ${\cal F}$ is the set of all bounded $1$-Lipschitz functions. This is also a constrained optimization due to the Lipschitz constraint on $f$ such that $f(x) - f(y) \le \|x-y\|$ for all $x, y\in {\cal X}$. Weight clipping \citep{arjovsky2017wasserstein} and gradient penalty \citep{gulrajani2017improved} have been used to satisfy the constraint of Lipschitz continuity. \cite{arjovsky2017wasserstein} used a clipping parameter $c$ to clamp each weight parameter to a fixed interval $[-c, c]$ after each gradient update is set. However, this method is very sensitive to the choice of clipping parameter $c$. Instead, \cite{gulrajani2017improved} introduced a gradient penalty, $\mathbb E_{\hat x} \big\{(\|\nabla_{\hat{x}} f(\hat{x})\|_2-1)^2\big\}$, in the loss function to enforce the Lipschitz constraint, where $\hat x$ is sampled uniformly along straight lines between pairs of points sampled from $P_X$ and $P_{G(Z)}$. This is motivated by the fact that the optimal critic contains straight lines with gradient norm $1$ connecting coupled points from $P_X$ and $P_{G(Z)}$. The experiment of \citep{arjovsky2017wasserstein} showed that the WGAN can avoid the problem of gradient vanishment. However, the WGAN does not produce meaningful encodings and many experiments still display the problem of mode collapse \citep{arjovsky2017wasserstein, gulrajani2017improved}. 

On the other hand, the Wasserstein Autoencoder (WAE) \citep{tolstikhin2018wasserstein}, after introducing an encoder $Q: {\cal X}\rightarrow {\cal Z}$ to approximate the conditional distribution of $Z$ given $X$, minimizes the reconstruction error $\inf_{Q\in {\cal Q}} \mathbb E_{X} \big \|X - G(Q(X)) \big\|$, where ${\cal Q}$ is a set of encoder mappings whose elements satisfies $P_{Q(X)}=P_Z$. The penalty, such as ${\cal D}(P_{Q(X)}, P_Z)$, is added to the objective to satisfy this constraint, where ${\cal D}$ is an arbitrary divergence between $P_{Q(X)}$ and $P_Z$. The WAE can produce meaningful encodings and has controlled reconstruction error. However, the WAE defines a generative model in an implicit way and does not model the generator through $G(Z)$ with $Z\sim P_Z$ directly.

To take the advantages of both the WGAN and WAE, we propose a new autoencoder GAN model, called the iWGAN, which defines the divergence between $P_X$ and $P_{G(Z)}$ by
\begin{equation}\label{equ:iwgan}
 \overline W_1(P_X, P_{G(Z)}) = \inf_{Q\in {\cal Q}} \sup_{f\in {\cal F}} \Big[\mathbb E_{X} \|X - G(Q(X))\| + \mathbb E_{X} \big\{f(G(Q(X)))\big\} - \mathbb E_{Z} \big\{f(G(Z))\big\}\Big].
\end{equation}
Our goal is to find the tuple $(G, Q, f)$ which minimizes $\overline W_1(P_X, P_{G(Z)})$. The motivation and explanation of this objective function are provided in Section \ref{sec:pd} in detail. The term $\|X- G(Q(X))\|$ can be treated as the autoencoder reconstruction error as well as a loss to match the distributions between $X$ and $G(Q(X))$. We note that the $L_1$-norm $\|\cdot\|_1$ has been used for the reconstruction term by the $\alpha$-GAN \citep{rosca2017variational} and CycleGAN \citep{zhu2017unpaired}. Another term $\mathbb E_{X\sim P_X}\{f(G(Q(X)))\} - \mathbb E_{Z\sim P_Z} \{f(G(Z))\}$ can be treated as a loss for the generator as well as a loss to match the distribution between $G(Q(X))$ and $G(Z)$. We emphasize that this term is different with the objective function of the WGAN in (\ref{eq:dual}). The properties of (\ref{equ:iwgan}) will be discussed in Theorem \ref{thm:main}, and the primal and dual explanation of (\ref{equ:iwgan}) will be presented in Section \ref{sec:pd}. 

Furthermore, it is challenging for practitioners to determine when to stop training GANs. Most of the GAN algorithms do not provide any explicit standard for the convergence of the model. However, the measure of convergence for the iWGAN becomes very natural and we use the duality gap as the measure. For a given tuple $(G, Q, f)$, the duality gap is defined as 
\begin{equation}\label{equ:dualgap}
\mbox{DualGap}(G, Q, f) = \sup_{\overline f\in {\cal F}} L(G, Q, \overline f) - \inf_{\overline G\in {\cal G}, \overline Q\in {\cal Q}} L(\overline G, \overline Q, f),
\end{equation}
where $L(G, Q, f)$ is
\begin{equation*}
L(G, Q, f) = \mathbb E_{X} \|X - G(Q(X))\| + \mathbb E_{X} \{f(G(Q(X)))\} - \mathbb E_{Z} \{ f(G(Z))\}.
\end{equation*}

In practice, the function spaces ${\cal G}$, ${\cal Q}$, and ${\cal F}$ are modeled by spaces containing deep neural networks with specific architectures.  The architecture hyperparameters usually include number of channels, number of layers, and width of each layer. The architectures for our numerical experiments are provided in the appendix. We assume that these network spaces are large enough to include the true encoder $Q^*$, generator $G^*$, and the optimal discriminator $f$ in (\ref{eq:dual}). This is not a strong assumption due to the universal approximation theorem of DNNs \citep{hornik1991approximation}.

\begin{theorem}\label{thm:main} 
(a). The iWGAN objective (\ref{equ:iwgan}) is equivalent to
\begin{align}\label{equ:iwgan2}
\overline W_1(P_X, P_{G(Z)}) = \inf_{Q\in\mathcal{Q}}\Big\{ W_1(P_X, P_{G(Q(X))}) + W_1(P_{G(Q(X))}, P_{G(Z)})\Big\}.
\end{align}
Therefore, $W_1(P_X, P_{G(Z)}) \le \overline W_1(P_X, P_{G(Z)})$.
If there exists a $Q^*\in {\cal Q}$ such that $Q^*(X)$ has the same distribution with $Z$, then $W_1(P_X, P_{G(Z)}) = \overline W_1(P_X, P_{G(Z)})$. \\
(b). Let $(\widetilde Q, \widetilde G, \widetilde f)$ be a fixed solution.
Then 
\[
\mbox{DualGap}(\tilde G, \tilde Q, \tilde f) \ge W_1(P_X, P_{\widetilde G(\widetilde Q(X))}) + W_1(P_{\widetilde G(\widetilde Q(X))}, P_{\widetilde G(Z)}).
\]
Moreover, if $\widetilde G$ outputs the same distribution as $X$ and $\widetilde Q$ outputs the same distribution as $Z$, both the duality gap and $\overline W_1(P_X, P_{\widetilde G(Z)})$ are zeros and $X = \widetilde G(\widetilde Q(X))$ for $X \sim P_X$.
\end{theorem}

According to Theorem \ref{thm:main}, the iWGAN objective is in general the upper bound of $W_1(P_X, P_{G(Z)})$. However, this upper bound is tight. When the space ${\cal Q}$ includes a special encoder $Q^*$ such that $Q^*(X)$ has the same distribution as $Z$, the iWGAN objective is exactly the same as $W_1(P_X, P_{G(Z)})$. Theorem \ref{thm:main} also provides an appealing property from a practical point of view. The values of both the duality gap and $\overline W_1(P_X, P_{\widetilde G(Z)})$ give us a natural criteria to justify the algorithm convergence.

\subsection{A Primal-Dual Explanation}\label{sec:pd}

We explain the iWGAN objective function (\ref{equ:iwgan}) from the view of primal and dual problems. Note that both the primal problem (\ref{eq:primal}) and the dual problem (\ref{eq:dual}) are constrained optimization problems. First, for the primal problem (\ref{eq:primal}), two constraints on $\pi$ are $\int \pi(x, z)dz - p_X(x)=0$ for all $x\in {\cal X}$, and $\int \pi(x, z)dz - p_Z(z)=0$ for all $z\in {\cal Z}$. Recall that the primal variable $f$ for the dual problem (\ref{eq:dual}) is also a dual variable for the primal problem (\ref{eq:primal}). From the Lagrange multiplier perspective, we can write the primal problem (\ref{eq:primal}) as 
\begin{align*}
&\resizebox{0.98\linewidth}{!}{$\inf_{\pi}\mathbb{E}_{\pi} \Big\|X- G(Z)\Big\| + \int_{x} f(x) \Big( p_X(x) - \int_z \pi(x, z) dz\Big) dx - \int_z f(G(z))\Big( p_Z(z) - \int_x \pi(x, z) dx\Big) dz$ }\\
=& \inf_{Q\in {\cal Q}}  \mathbb E_{X} \Big\{ \|X - G(Q(X)) \| + f(G(Q(X)))\Big\}  - \mathbb E_Z \big\{ f(G(Z)) \big\},
\end{align*}
where we use the encoder $Q$ to approximate the conditional distribution of $Z$ given $X$, and the Lagrange multipliers for two constraints are $f(x)$ and $-f(G(z))$ respectively. Second, for the dual problem (\ref{eq:dual}), the $1$-Lipschitz constraint on $f$ is $f(x) - f(G(z)) \le \|x - G(z)\|$ for all $x\in {\cal X}$ and $z\in {\cal Z}$. Recall that the primal variable $\pi$ for the primal problem (\ref{eq:primal}) is also a dual variable for the dual problem (\ref{eq:dual}). Similarly, we can write the dual problem (\ref{eq:dual}) as
\begin{align*}
&\sup_{f\in {\cal F}} \mathbb E_{X} \big\{ f(X)\big\} - \mathbb E_{Z} \big\{ f(G(Z))\big\} - \int_{{\cal X}\times\cal Z} \pi(x, z) \Big(f(x)-f(G(z))-\|x-G(z)\|\Big) dxdz\\
=& \sup_{f\in {\cal F}}  \mathbb E_{X} \Big\{ \|X - G(Q(X)) \| + f(G(Q(X))) \Big\} - \mathbb E_Z \big\{ f(G(Z)) \big\},
\end{align*} 
where the Lagrange multiplier for the $1$-Lipschitz constraint is $\pi(x, z)$. When we solve primal and dual problems iteratively, this turns out to be exactly the same as our iWGAN algorithm. 

In addition, the optimal value of the primal problem (\ref{eq:primal}) satisfies
\begin{align*}
\inf_{Q\in {\cal Q}} \sup_{f\in {\cal F}} \mathbb E_{X} \Big\{ \|X - G(Q(X)) \| + f(G(Q(X)))\Big\}  - \mathbb E_Z \big\{ f(G(Z)) \big\},
\end{align*}
and the optimal value of the dual problem (\ref{eq:dual}) satisfies
\begin{align*}
\sup_{f\in {\cal F}}\inf_{Q\in {\cal Q}}  \mathbb E_{X} \Big\{ \|X - G(Q(X)) \| + f(G(Q(X))) \Big\} - \mathbb E_Z \big\{ f(G(Z)) \big\}.
\end{align*} 
The difference between the optimal primal and dual values is exactly the duality gap in (\ref{equ:dualgap}).

\subsection{Extension to f-GANs}\label{sec:fgan}

This framework can be easily extended to other types of GANs. Assume that $\cal F$ is the 1-Lipschitz function class. We extend the iWGAN framework to the inferential f-GAN (ifGAN) framework. Define the ifGAN objective function as follows: 
\begin{equation}\label{equ:ifgan}
\resizebox{0.93\textwidth}{!}{
 $\overline W_{1, h}(P_X, P_{G(Z)}) = \inf_{Q\in {\cal Q}} \sup_{f\in {\cal F}} \Big[\mathbb E_{X} \|X - G(Q(X))\| + \mathbb E_{X} \big\{f(G(Q(X)))\big\} - \mathbb E_{Z} \big\{h^{*}(f(G(Z)))\big\}\Big]$.}
\end{equation}
Following this definition, we have
\begin{align*}
\overline W_{1, h}(P_X,P_{G(Z)}) = \inf_{Q\in{\cal Q}}\Big\{W_{1}(P_X,P_{G(Q(X))}) +\text{GAN}_h(P_{G(Q(X))}, P_{G(Z)})\Big\}.
\end{align*}
We show  $\text{GAN}_h(P_X,P_{G(Z)})\le\overline W_{1,h}(P_X, P_{G(Z)})$. This is because
\begin{align*}
 &   \text{GAN}_h(P_X, P_{G(Z)}) =  \sup_{f\in\cal F}\mathbb E_{X}\left\{f(X)\right\} - \mathbb E_{Z}\left\{h^{*}(f(G(Z))\right\}\\
    \le & \inf_{Q\in\cal Q}\Big[\sup_{f\in \cal F}\mathbb E_{X}\left\{f(X)\right\}  - \mathbb E_{X}\left\{f(G(Q(X)))\right\} +\sup_{f\in \cal F}\mathbb E_{X}\left\{f(G(Q(X)))\right\}  - \mathbb E_{Z}\left\{h^{*}(f(G(Z))\right\}\Big]\\
    =& \overline W_{1,h}(P_X,P_{G(Z)}).
\end{align*}
This indicates that the ifGAN objective (\ref{equ:ifgan}) is an upper bound of the f-GAN objective.

\section{Generalization Error Bound and the Algorithm}

Suppose that we observe $n$ samples $\{x_1, \ldots, x_n\}$. In practice, we minimize the empirical version, denoted by $\widehat{\overline W}_1(P_X, P_{G(Z)})$, of $\overline W_1(P_X, P_{G(Z)})$ to learn both the encoder and the generator, where,
\begin{equation}\label{equ:emw}
\widehat{\overline W}_1(P_X, P_{G(Z)}) = \inf_{Q\in {\cal Q}} \sup_{f\in {\cal F}} \Big[\hat{\mathbb E}_{obs} \|x - G(Q(x))\| + \hat{\mathbb E}_{obs} \big\{f(G(Q(x)))\big\} - \hat{\mathbb E}_{z} \big\{ f(G(z))\big\}\Big].
\end{equation}
Here $\hat{\mathbb{E}}_{obs}\{\cdot\}$ denotes the empirical average on the observed data $\{x_{i}\}$ and $\hat{\mathbb E}_{z}$ denotes the empirical average on a random sample of standard normal random variables. Before we present the details of the algorithm, we first establish the generalization error bound for the iWGAN in this section. 

In the context of supervised learning, generalization error is defined as the gap between the empirical risk and the expected risk. The empirical risk is corresponding to the training error, and the expected risk is corresponding to the testing error. Mathematically, the difference between the expected risk and the empirical risk, i.e. the generalization error, is a measure of how accurately an algorithm is able to predict outcome values for previously unseen data. However, in the context of GANs,  neither the training error nor the test error is well defined. But we can define the generalization error in a similar way. Explicitly, we define the ``training error" as $\widehat{\overline W}_1(P_X, P_{G(Z)})$ in (\ref{equ:emw}), which is minimized based on observed samples. Define the ``test error" as  $W_1(P_X, P_{G(Z)})$ in (\ref{eq:primal}), which is the true 1-Wasserstein distance between $P_X$ and $P_{G(Z)}$. The generalization error for the iWGAN is defined as the gap between these two ``errors". In other words, for an iWGAN model with the parameter $(G, Q, f)$, the generalization error is defined as $\widehat{\overline W}_1(P_X, P_{G(Z)}) - W_1(P_X, P_{G(Z)})$. For discussions of generalization performance of classical GANs, see \cite{arora2017generalization} and \cite{jiang2018computation}.

\begin{theorem}\label{thm:WGANgeneralize}
Given a generator $G\in\mathcal{G}$, and $n$ samples $(x_1,\ldots,x_n)$ from $\mathcal{X} = \{x:\|x\|\le B\}$, with probability at least $1-\delta$ for any $\delta\in(0,1)$, we have
\begin{align}
    W_1(&P_X, P_{G(Z)}) \le \widehat{\overline{W}}_{1}(P_X, P_{G(Z)})+2\widehat{\mathfrak{R}}_n(\mathcal{F})+3B\sqrt{\frac{2}{n}\log\left(\frac{2}{\delta}\right)},
\end{align}
where $\widehat{\mathfrak{R}}_n(\mathcal{F})=\E_{\epsilon}\left\{\sup_{f\in \mathcal{F}}n^{-1}\sum_{i=1}^n\epsilon_if(x_i)\right\}$ is the empirical Rademacher complexity of the 1-Lipschitz function set $\mathcal{F}$, in which $\epsilon_i$ is the Rademacher variable.
\end{theorem}

For a fixed generator $G$, Theorem \ref{thm:WGANgeneralize} holds uniformly  for any discriminator $f\in\mathcal{F}$. It indicates that the 1-Wasserstein distance between $P_X$ and $P_{G(Z)}$ can be dominantly upper bounded by the empirical $\widehat{\overline{W}}_1(P_X,P_{G(Z)})$ and Rademacher complexity of $\mathcal{F}$. Since $\widehat{\overline W}_1(P_X, P_{G(Z)}) \le \widehat W_1(P_X, P_{G(Q(X))}) + \widehat W_1(P_{G(Q(X))}, P_{G(Z)})$ for any $Q\in {\cal Q}$, the capacity of $\mathcal{Q}$ determines the value of $\widehat{\overline W}_1(P_X, P_{G(Z)})$. In learning theory, Rademacher complexity, named after Hans Rademacher, measures richness of a class of real-valued functions with respect to a probability distribution. There are several existing results on the empirical Rademacher complexity of neural networks. For example, when $\mathcal{F}$ is a set of 1-Lipschitz neural networks, we can apply the conclusion from  \cite{bartlett2017spectrally} to $\widehat{\mathfrak{R}}_n(\mathcal{F})$, which produces an upper bound scaling as $\mathcal{O}(B\sqrt{L^3/n})$. Here $L$ denotes the depth of network $f\in\mathcal{F}$. Similar upper bound with an order of $\mathcal{O}(B\sqrt{Ld^2/n})$ can be obtained by utilizing the results from \cite{li2018tighter}, where $d$ is the width of the network.

\begin{algorithm}[H]
    \small
    \caption{The training algorithm of iWGAN}\label{algo}
    \begin{algorithmic}[1]
        \Require The regularization coefficients $\lambda_1$ and $\lambda_2$, tolerance for duality gap $\epsilon_1$, tolerance for loss $\epsilon_2$, and running steps $n$
        \State Initialization $(G^0, Q^0, f^0)$  
        \While{$\mbox{DualGap}(G^i, Q^i, f^i) > \epsilon_1$ or $L(G^i, Q^i, f^i) > \epsilon_2$}
            \For{$t = 1$, ..., $n$} 
                \State Sample real data $\{x^i_k\}_{k=1}^n \sim {P}_X$, latent variable $\{z^i_k\}_{k=1}^n \sim {P}_Z$ and $\{\epsilon_k \}_{k=1}^n \sim U[0, 1]$
                \State Set $\hat{x}^i_k \leftarrow \epsilon_k x^i_k + (1 - \epsilon_k) G^i(z^i_k)$, $i=1, ..., n$ for the calculation of gradient penalty
                \State Calculate: $L^i = L(G^i, Q^i, f^i)$, $J_1(f^i) = (\|\nabla_{\hat{x}^i} f^i(\hat{x}^i)\|_2 - 1)^2$, and
                    \begin{align*}
                    -\nabla_f L^i &= \nabla_f \Big[\dfrac{1}{n}\sum_{k=1}^n\Big(f^i(G^i(z_k^i)) - f^i(G^i(Q^i(x_k^i))) + \lambda_1 J_1(f^i)\Big)\Big]
                    \end{align*}
                \State Update $f$ by Adam: $f^{i+1} \leftarrow f^i + Adam(-\nabla_f L^i)$
            \EndFor
            \For{$t = 1$, ..., $n$} 
                \State Sample real data $\{x^i_k\}_{k=1}^n \sim {P}_X$, latent variable $\{z^i_k\}_{k=1}^n \sim {P}_Z$ 
                \State Calculate: $L'^i = L(G^i, Q^i, f^{i+1})$, $J_2(Q^i)$, and
                \begin{align*}
                    \nabla_{G, Q} L'^i &=\nabla_{G, Q}\Big[ \dfrac{1}{n} \sum_{k=1}^n \Big(\|x_k^i- G^i(Q^i(x_k^i))\| + f^{i+1}(G^i(Q^i(x_k^i))) - f^{i+1}(G^i(z_k^i)) + \lambda_2 J_2(Q^i)\Big)\Big]
                \end{align*}
                \State Update $G$, $Q$ by Adam: $(G^{i+1}, Q^{i+1}) \leftarrow (G^{i}, Q^{i}) + Adam(\nabla_{G, Q} L'^i)$ 
            \EndFor
        
        \State DualGap$(G^{i+1}, Q^{i+1}, f^{i+1}) = L(G^i, Q^i, f^{i+1}) - L(G^{i+1}, Q^{i+1}, f^{i+1})$
        \State $i \leftarrow i+1$
        \EndWhile
    \end{algorithmic}
\end{algorithm}

Next, we introduce the details of the algorithm. Our target is to solve the following optimization problem:
\begin{equation}\label{equ:obj2}
\resizebox{0.93\textwidth}{!}{    
$\min\limits_{G\in {\cal G}, Q\in {\cal Q}} \max\limits_{f\in {\cal F}} \Big[ \hat{\mathbb E}_{obs} \|x - G(Q(x))\| + \hat{\mathbb E}_{obs} \big\{f(G(Q(x)))\big\} - \hat{\mathbb E}_{z} \big\{f(G(z))\big\} 
    - \lambda_1 J_1(f) + \lambda_2 J_2(Q)\Big]$, }
\end{equation}
where $J_1(f)$ and $J_2(Q)$ are regularization terms for $f$ and $Q$ respectively. We approximate $G, Q, f$ by three neural networks with pre-specified architectures. 

Since $f$ is assumed to be 1-Lipschitz,  we adopt the gradient penalty defined as $J_1(f) = \mathbb E_{\hat x}\big\{(\mathbb \|\nabla_{\hat x} f(\hat x)\|_2 - 1)^2\big\}$ in \citep{gulrajani2017improved} to enforce the 1-Lipschitz constraint on $f\in {\cal F}$. Furthermore, since we expect $Q(X)$ follows approximately standard normal, we use the maximum mean discrepancy (MMD) penalty \citep{gretton2012kernel}, denoted by $J_2(Q) = \mbox{MMD}_k(P_{Q(X)}, P_Z)$, to enforce  $Q(X)$ to converge to $P_Z$. In particular,
\[
J_2(Q) =\dfrac{1}{n(n-1)} \sum_{l\neq j} k(z^i_l, z^i_j) + \dfrac{1}{n(n-1)} \sum_{l\neq j} k(Q(x^i_l), Q(x^i_j)) - \dfrac{2}{n^2} \sum_{l, j} k(z^i_l, Q(x^i_j)),
\]
where $k$ is set to be the Gaussian radial kernel function $k(x, y) = \exp(\frac{-\|x-y\|^2}{2})$.

We have adopted the stochastic gradient descent algorithm called the ADAM \citep{kingma2014method} to estimate the unknown parameters in neural networks. The ADAM is an algorithm for first-order gradient-based optimization of stochastic objection functions, based on adaptive estimates of lower-order moments. Given the current tuple $(G^i, Q^i, f^i)$ at the $i$th iteration, we sample a batch of observations $\{x^i_k\}_{k=1}^n \sim {P}_X$, latent variable $\{z^i_k\}_{k=1}^n \sim {P}_Z$, and $\{\epsilon_k \}_{k=1}^n \sim U[0, 1]$. Then we construct $\hat{x}^i_k \leftarrow \epsilon_k x^i_k + (1 - \epsilon_k) G^i(z^i_k)$, $i=1, \ldots, n$, for computing the gradient penalty. Let $L^i = L(G^i, Q^i, f^i)$ and $J_1(f^i) = (\|\nabla_{\hat{x}^i} f^i(\hat{x}^i)\|_2 - 1)^2$. We can evaluate the gradient with respect to the parameters in $f$, which is denoted by
\begin{align*}
    -\nabla_f L^i = \nabla_f \Big[\dfrac{1}{n}\sum_{k=1}^n\Big(f^i(G^i(z_k^i)) - f^i(G^i(Q^i(x_k^i))) + \lambda_1 J_1(f^i)\Big)\Big].
\end{align*}
Then we can update $f^i$ by the ADAM using this gradient. Similarly, we can evaluate the gradient with respect to the parameters in $G$ and $Q$,   which is denoted by
\[
\nabla_{G, Q} L^i =\nabla_{G, Q}\Big[ \dfrac{1}{n} \sum_{k=1}^n \Big(\|x_k^i- G^i(Q^i(x_k^i))\| + f^{i+1}(G^i(Q^i(x_k^i))) - f^{i+1}(G^i(z_k^i)) + \lambda_2 J_2(Q^i)\Big)\Big].
\]
Then we can update $(G^i, Q^i)$ by the ADAM using this gradient. The stopping criteria are both the DualGap$(G^i, Q^i, f^i)$ in (\ref{equ:dualgap}) and the objective function $L(G^i, Q^i, f^i)$ are less than pre-specified error tolerances $\epsilon_1$ and $\epsilon_2$, respectively. Specifically, based on the definition of the duality gap in (\ref{equ:dualgap}), we approximate DualGap$(G^i, Q^i, f^i)$ by the difference between $L(G^i, Q^i, f^{i+1})$ and The optimization (\ref{equ:obj2}) consists of two tuning parameters $\lambda_1$ and $\lambda_2$. We pre-specify some values for $\lambda_1$ and $\lambda_2$ and select the optimal tuning parameters by grid search using cross validation.
The details of the algorithm are presented in Algorithm \ref{algo}.

\section{Probabilistic Interpretation and the MLE }\label{sec:PI}

The iWGAN has proposed an efficient framework to stably and automatically estimate both the encoder and the generator. In this section, we provide a probabilistic interpretation of the iWGAN under the framework of maximum likelihood estimation.

Maximum likelihood estimator (MLE) is a fundamental statistical framework for learning models from data. However, for complex models, MLE can be computationally prohibitive due to the intractable normalization constant. MCMC has been used to approximate the intractable likelihood function but do not work efficiently in practice, since running MCMC till convergence to obtain a sample can be computationally expensive. For example, to reduce the computational complexity, \citet{hinton2002training} proposed a simple and fast algorithm, called the contrastive divergence (CD). The basic idea of CD is to truncate MCMC at the $k$-th step, where $k$ is a fixed integer as small as one. The simplicity and computational efficiency of CD makes it widely used in many popular energy-based models. However, the success of CD also raised a lot of questions regarding its convergence property. Both theoretical and empirical results show that CD in general does not converge to a local minimum of the likelihood function \citep{carreira2005contrastive, qiu20}, and diverges even in some simple models \citep{schulz2010investigating,fischer2010empirical}. The iWGAN can be treated as an adaptive method for the MLE training, which not only provides computational advantages but also allows us to generate more realistic-looking images. Furthermore, this probabilistic interpretation enables other novel applications such as image quality checking and outlier detection. 

Let $X$ denote the image. Define the density of $X$ by an energy-based model based on an autoencoder \citep{gu01, zhao2016energy, berthelot2017began}:
\begin{equation}\label{equ:ebm}
p(x|\theta) = \exp\big(-\big\|x - G_{\theta}(Q_{\theta}(x))\big\| - V(\theta)\big),
\end{equation}
where
\[ V(\theta) = \log \int \exp(-\big\|x - G_{\theta}(Q_{\theta}(x))\big\|) dx,
\]
and $\theta\in \Theta$ is the unknown parameter and $V(\theta)$ is the log normalization constant. The major difficulty for the likelihood inference is due to the intractable function $V(\theta)$. Suppose that we have the observed data $\{x_{i}: i=1,\ldots, n\}$. The log-likelihood function of $\theta \in \Theta$ is $\ell(\theta) = n^{-1}\sum_{i=1}^n \log~p(x_{i}| \theta)$, whose gradient is
\begin{equation}\label{equ:grad}
    \nabla_{\theta} \ell(\theta) =  -\hat{\mathbb{E}}_{obs}\big\{\partial_{\theta} \big\|x - G_\theta(Q_\theta(x))\big\|\big\} + \mathbb{E}_{\theta}\big\{\partial_{\theta}\big\|x - G_\theta(Q_\theta(x))\big\|\big\},
\end{equation}
where $\hat{\mathbb{E}}_{obs}[\cdot]$ denotes the empirical average on the observed data $\{x_{i}\}$ and $\mathbb{E}_\theta[\cdot]$ denotes the expectation under model $p(x|\theta)$. The key computational obstacle lies in the approximations of the model expectation $\mathbb{E}_\theta[\cdot]$.

To address this problem, we can rewrite the log-likelihood function by introducing a variational distribution $q(x)$. This leads to
\begin{align}\label{equ:upper}
    \hat{\mathbb E}_{obs} \log p(x|\theta) &= - \hat{\mathbb E}_{obs} \|x - G_\theta(Q_\theta(x))\| - V(\theta) \nonumber\\
    &=- \hat{\mathbb E}_{obs} \|x - G_\theta(Q_\theta(x))\|  - \log \int q(x) {e^{-\|x - G_\theta(Q_\theta(x))\|}\over q(x)}dx \nonumber \\
    &\le - \hat{\mathbb E}_{obs} \|x - G_\theta(Q_\theta(x))\|-\int q(x) \log{e^{-\|x - G_\theta(Q_\theta(x))\|}\over q(x)}dx \nonumber \\
    & = - \hat{\mathbb E}_{obs} \|x - G_\theta(Q_\theta(x))\| + \mathbb E_{q(x)} \|x - G_\theta(Q_\theta(x))\| - H(q),
\end{align}
where $H(q)=-\int q\log q$ denotes the entropy of $q(x)$ and the inequality is due to Jensens's inequality. Equation (\ref{equ:upper}) provides an upper bound for the log-likelihood function. We expect to choose $q(x)$ so that (\ref{equ:upper}) is closer to the log-likelihood function, and then maximize (\ref{equ:upper}) as a surrogate for maximizing log-likelihood. We choose $q(x) = p(x|\tilde\theta)$ and define the surrogate log-likelihood as
\begin{equation}\label{equ:surr}
{\cal L}(\theta; \tilde\theta) = - \hat{\mathbb E}_{obs} \|x - G_\theta(Q_\theta(x))\| + \mathbb E_{\tilde\theta} \|x - G_\theta(Q_\theta(x))\| - H(p(x|\tilde\theta)).
\end{equation}

\begin{theorem}\label{thm:up}
(a). For any $\theta, \tilde\theta \in \Theta$, we have $\ell(\theta) \le {\cal L}(\theta; \tilde\theta)$. In addition, $\ell(\theta) = {\cal L}(\theta; \theta)$.\\
(b). Consider the following algorithm and the $(t+1)$th iteration is obtained by $\theta^{(t+1)} = \arg\max_{\theta\in \Theta} {\cal L}(\theta; \theta^{(t)})$, for $t=0, 1, \cdots$. If $\theta^{(t)}\rightarrow \hat\theta$ as $t\rightarrow \infty$, then $\hat\theta$ is the MLE.
\end{theorem}

Theorem \ref{thm:up} shows that, if we maximize the surrogate log-likelihood function and the algorithm converges, the solution is exactly the same as the MLE. The additional identity $\ell(\theta) = {\cal L}(\theta; \theta)$ is the key to our algorithm to obtain the MLE, which is different with the ELBO in VAEs. The ELBO is in general not a tight lower bound of the log-likelihood function.

In terms of training, by Theorem 5.10 of \cite{villani2008optimal}, for any random variables $X$ and $Y$, there exists an optimal $f^*$ such that 
\begin{equation}\label{equ:coup}
    \mathbb P_{\pi}\big( f^*(Y) - f^*(X) = \|Y - X\| \big)= 1  
\end{equation}
for the optimal coupling $\pi$ which is the joint distribution of $X$ and $Y$. Therefore, there exists a $f^*$ such that
\begin{equation}\label{equ:tmp11}
    f^*(X)-f^*(G_{\theta}(Q_{\theta}(X)))=\|X-G_{\theta}(Q_{\theta}(X))\|
\end{equation}
with probability one. Because $f^*$ needs to be learned as well, we approximate $f^*$ by a neural network $f_\eta$  with an unknown parameter $\eta$. This amounts to using the following max-min objective
\begin{equation}\label{equ:upper2}
    \max_\theta \min_\eta - \hat{\mathbb E}_{obs} \|x - G_\theta(Q_\theta(x))\| + \mathbb E_{\theta^{(t)}} f_\eta(x)- \mathbb E_{\theta^{(t)}} f_\eta(G_{\theta}(Q_{\theta}(x))).
\end{equation}
Note that, for the gradient update, the expectation in (\ref{equ:upper2}) is taken under the current estimated $\theta^{(t)}$. Since we require $G_\theta$ to be a good generator and the distributions of $G_\theta(z)$ is close the distribution $p(x|\theta^{(t)})$, we replace $\mathbb E_{\theta^{(t)}} f_\eta(x)$ by $\mathbb E_zf_\eta(G_\theta(z)))$. Since an additional regularization is added to enforce $Q_\theta(X)$ to follow a normal distribution, we use the expectation under the data distribution to replace the second expectation of (\ref{equ:upper2}). This yields a gradient update for $\theta$ of form $\theta\leftarrow \theta + \epsilon \hat\nabla_\theta \ell(\theta)$, where
\begin{equation}\label{equ:grad2}
    \hat\nabla_\theta \ell(\theta) = - \hat{\mathbb{E}}_{obs}\big\{\partial_{\theta} \big\|x - G_\theta(Q_\theta(x))\big\|\big\} +\big\{ {\mathbb{E}}_{z}  \partial_{\theta}f_\eta(G_\theta(z)) -\mathbb E_{x}\partial_{\theta}f_\eta(G_\theta(Q_\theta(x))\big\}. 
\end{equation}
A gradient update for $\eta$ is given by 
\begin{equation}\label{equ:grad3}
    \eta \leftarrow \eta + \epsilon  ~ \Big\{{\mathbb{E}}_{z} \partial_{\eta}f_\eta(G_\theta(z)) -\mathbb E_x\partial_{\eta}f_\eta(G_\theta(Q_\theta(x))\Big\}.
\end{equation}
The above iterative updating process is exactly the same as in Algorithm \ref{algo}. Therefore, the training of the iWGAN is to seek the MLE. This probabilistic interpretation provides a novel alternative method to tackle problems with the intractable normalization constant in latent variable models. The MLE gradient update of $p(x|\theta)$ decreases the energy of the training data and increases the dual objective. Compare with original GANs or WGANs, our method gives much faster convergence and simultaneously provides a higher quality generated images. 

The probabilistic modeling opens a door for many interesting applications. Next, we present a completely new approach for determining a highest density region (HDR) estimate for the distribution of $X$. What makes HDR distinct from other statistical methods is that it finds the smallest region, denoted by $U(\alpha)$, in the high dimensional space with a given probability coverage $1-\alpha$, i.e., $\mathbb P(X \in U(\alpha)) = 1-\alpha$. We can use $U(\alpha)$ to assess each individual sample quality. Note that commonly used inception scores (IS) and Fréchet inception distances (FID) are to measure the whole sample quality, not at the individual sample level. More introductions of IS and FID are given in Appendix G.  Let $\hat\theta$ be the MLE. The density ratio at $x_1$ and $x_2$  is 
\[ \frac{p(x_1|\hat\theta)}{p(x_2|\hat\theta)}= \exp\big\{-(\|x_1 - G_{\hat\theta}(Q_{\hat\theta}(x_1))\|-\|x_2 - G_{\hat\theta}(Q_{\hat\theta}(x_2))\|)\big\}.
\]
The smaller the reconstruction error is, the larger the density value is. We can define the HDR for $x$ through the HDR for the reconstruction error $e_x: = \|x - G_{\hat\theta}(Q_{\hat\theta}(x))\|$, which is simple because it is a one-dimensional problem. Let $\tilde U(\alpha)$ be the HDR for $e_x$. Then, $U(\alpha) = \{x: e_x \in \tilde U(\alpha) \}$. Here $Q_{\hat\theta}(U(\alpha))$ defines the corresponding region in the latent space, which can be used to generate better quality samples.

\section{Experimental Results}

The goal of our numerical experiments is to demonstrate that the iWGAN can achieve the following three objectives simultaneously: high-quality generative samples, meaningful latent codes, and small reconstruction errors. We also compare the iWGAN with other well-known GAN models such as the Wasserstein GAN with gradient penalty (WGAN-GP) \citep{gulrajani2017improved}, the Wasserstein Autoencoder (WAE) \citep{tolstikhin2018wasserstein}, the Adversarial Learned Inference (ALI) \citep{dumoulin2017adversarially}, and the CycleGAN \citep{zhu2017unpaired} to illustrate a competitive and stable performance for benchmark datasets.

\subsection{Mixture of Gaussians}

\begin{figure}[t]
    \centering
    \begin{subfigure}{0.28\linewidth}
    \centering
    \includegraphics[width=0.9\linewidth]{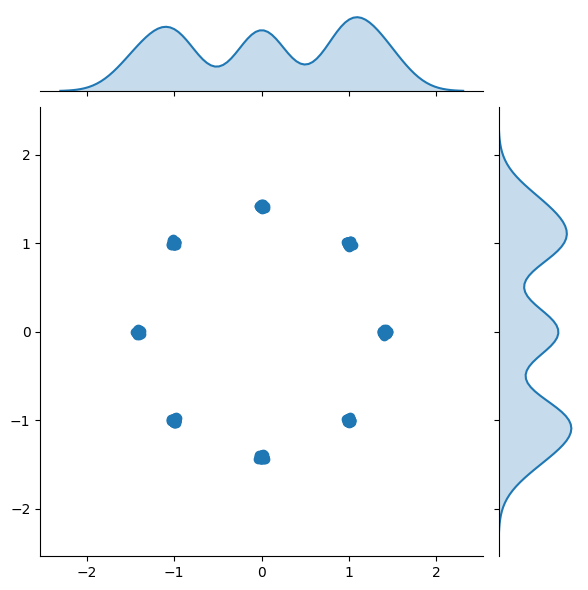}
    \caption{RING}
    \label{fig:ring}
    \end{subfigure}
    \hfill
    \begin{subfigure}{0.28\linewidth}
    \centering
    \includegraphics[width=0.9\linewidth]{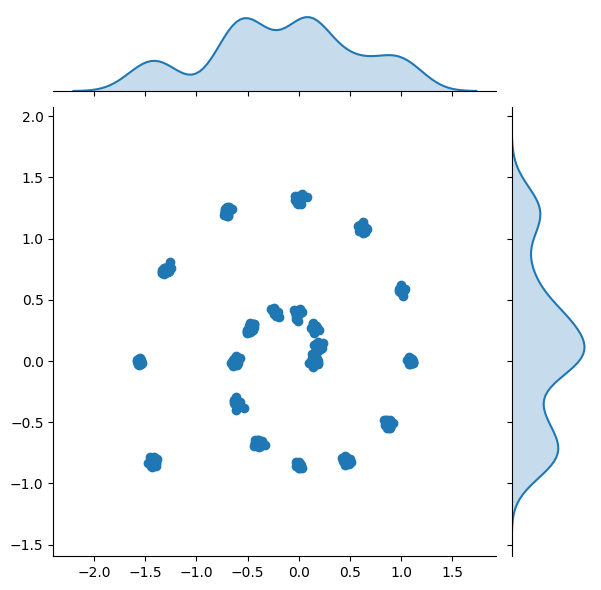}
    \caption{Swiss Roll}
    \label{fig:sr}
    \end{subfigure}
    \hfill
    \begin{subfigure}{0.28\linewidth}
    \centering
    \includegraphics[width=0.9\linewidth]{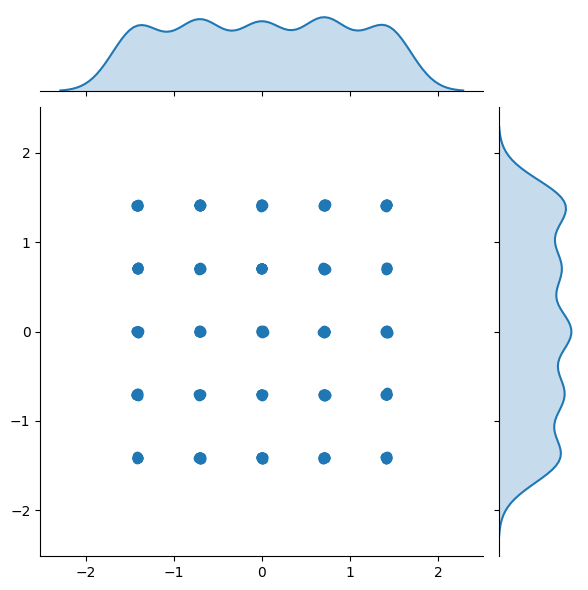}
    \caption{GRID}
    \label{fig:grid}
    \end{subfigure}
    \caption{Samples of mixture of Gaussians}
    \label{fig:real}
\end{figure}

We first train our iWGAN model on three datasets from the mixture of Gaussians with an increasing difficulty shown in the Figure \ref{fig:real}: a). RING: a mixture of 8 Gaussians with means $\{(2 \cdot \cos{\dfrac{2 \pi i}{8}}, 2 \cdot \sin{\dfrac{2 \pi i}{8}})| i = 0, \dots 7\}$ and standard deviation $0.02$, b). SPIRAL: a mixture of 20 Gaussians with means $\{(0.1+0.1\cdot \dfrac{2\pi}{20}\cdot \cos{\dfrac{2\pi i}{20}}, 0.1+0.1\cdot \dfrac{2\pi}{20}\cdot \sin{ \dfrac{2\pi i}{20}})|i = 0, \dots, 19\}$ and standard deviation $0.02$ and c). GRID: a mixture of 25 Gaussians with means $\{(2\cdot i, 2\cdot j)|i = -2, -1, \dots, 2, j = -2, -1, \dots, 2\}$ and standard deviation $0.02$. As the true data distributions are known, this setting allows for tracking of convergence and mode dropping.

\begin{figure*}[t]
    \centering
    \includegraphics[width=1.\textwidth, height=2.25in]{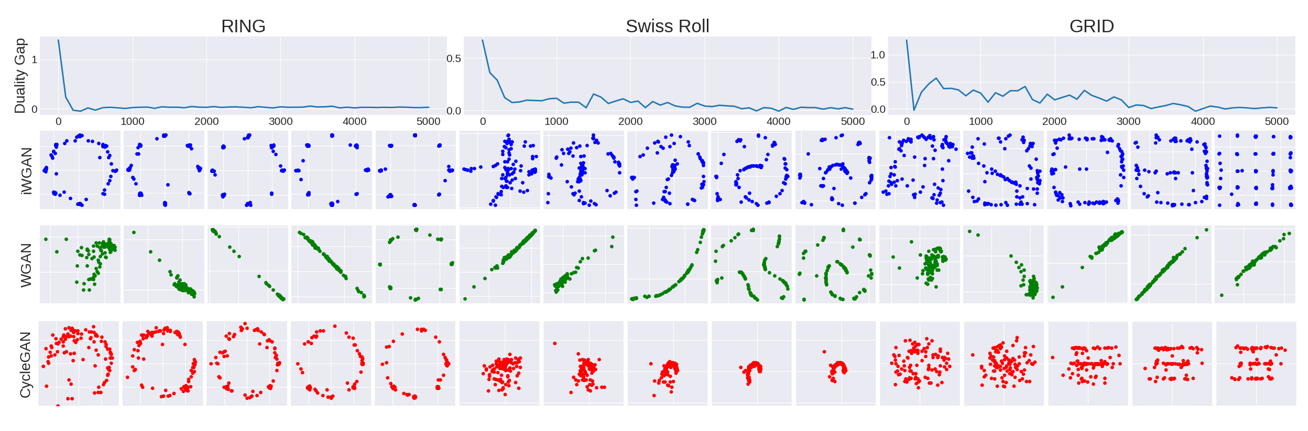}
    \caption{Duality gap, generated samples from the iWGAN, WGAN-GP, and CycleGAN on mixture of Gaussians at 1000, 2000, ..., 5000 epochs. First row: The duality gaps of the iWGAN in 3 experiments which converge to 0 very fast; Second row: Generated samples of the iWGAN. The iWGAN has successfully generated samples following all target distributions. Third row: Generated samples of the WGAN-GP. The WGAN-GP has successfully generated samples for Ring and Swiss Roll, but has failed for Grid. Fourth row: Generated samples of the CycleGAN. The CycleGAN has failed on all 3 distributions and experienced the mode collapse problem.}
    \label{fig:losses}
\end{figure*}

\noindent{\bf Duality gap and convergence.} We illustrate that as the duality gap converges to $0$, our model converges to the generated samples from the true distribution. We keep track of the generated samples using $G(z)$ and record the duality gap at each iteration to check the corresponding generated samples. We compare our method with the WGAN-GP and CycleGAN in Figure \ref{fig:losses}. All methods adopt the same structure, learning rate, number of critical steps, and other hyper-parameters. 

Figure \ref{fig:losses} shows that the iWGAN converges quickly in terms of both the duality gap and the true distributions learning. Duality gap has also been a good indicator of whether the model has generated the desired distribution. When comparing with the WGAN model, the iWGAN surpasses the performance of the WGAN-GP at very early stage and avoids the appearance of mode collapse. We have further tested the CycleGAN on these distributions.  The CycleGAN objective function is the sum of  two parts. The first part includes two vanilla GAN objectives \citep{goodfellow2014generative} to differentiate between $X$ and $G(Z)$, and $Z$ and $Q(X)$. The second part is the cycle consistency loss given by $ \mathbb E_{Z} \big\|Z - Q(G(Z)) \|_1 + \mathbb E_{X} \big\|X - G(Q(X)) \|_1$, where $\|\cdot\|_1$ is the $L_1$-norm of a vector. Unfortunately, Figure \ref{fig:losses} shows that  the CycleGAN fails on all three distributions and experienced the mode collapse problem.

\begin{figure*}[!ht]
    \centering
    \begin{subfigure}{0.3\textwidth}
        \centering
        \includegraphics[width=1.0\linewidth, height=2.25in]{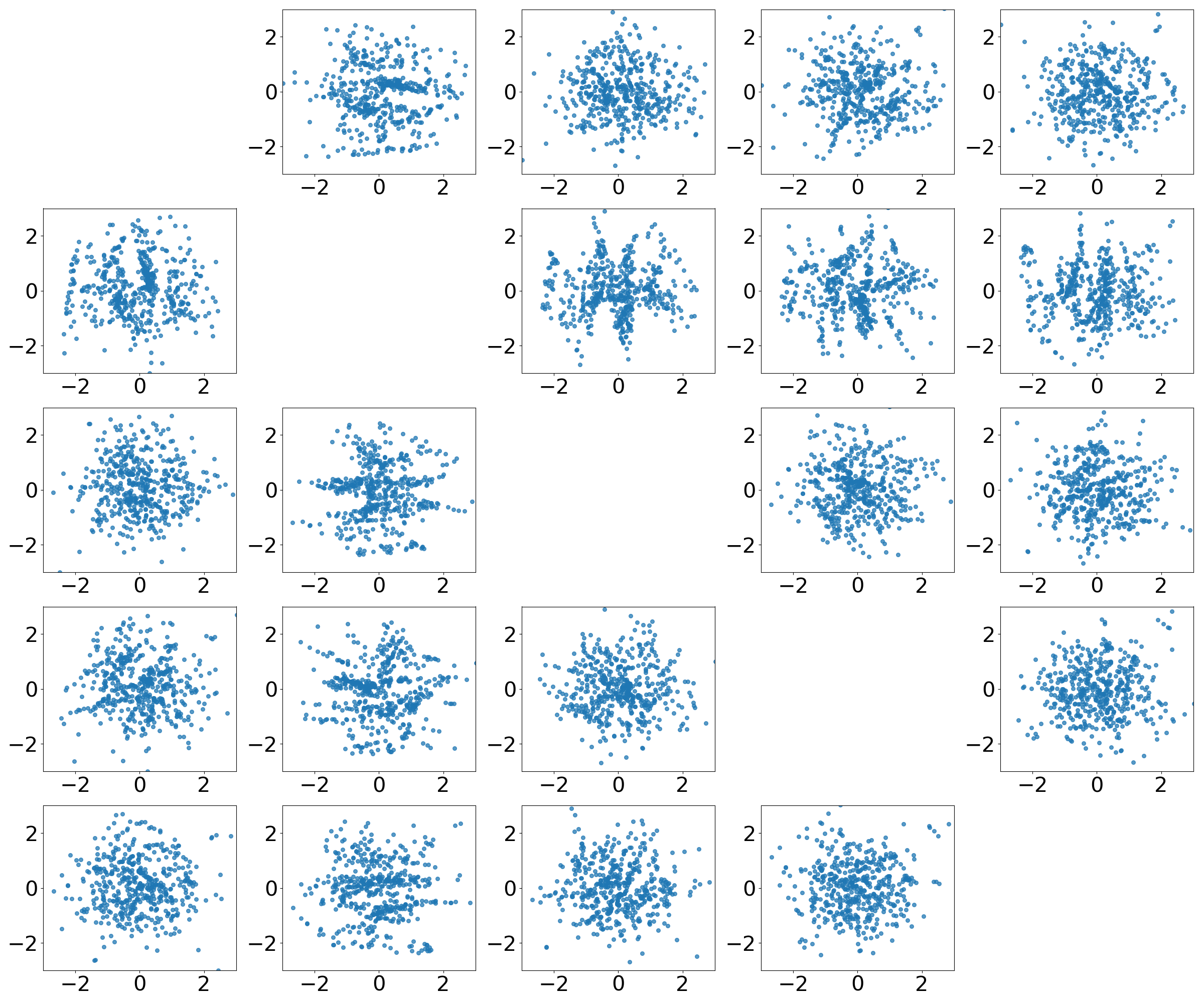}
        \caption{RING}
        \label{fig:latent_ring}
    \end{subfigure}
    \hfill
    \begin{subfigure}{0.3\textwidth}
        \centering
        \includegraphics[width=1.0\linewidth, height=2.25in]{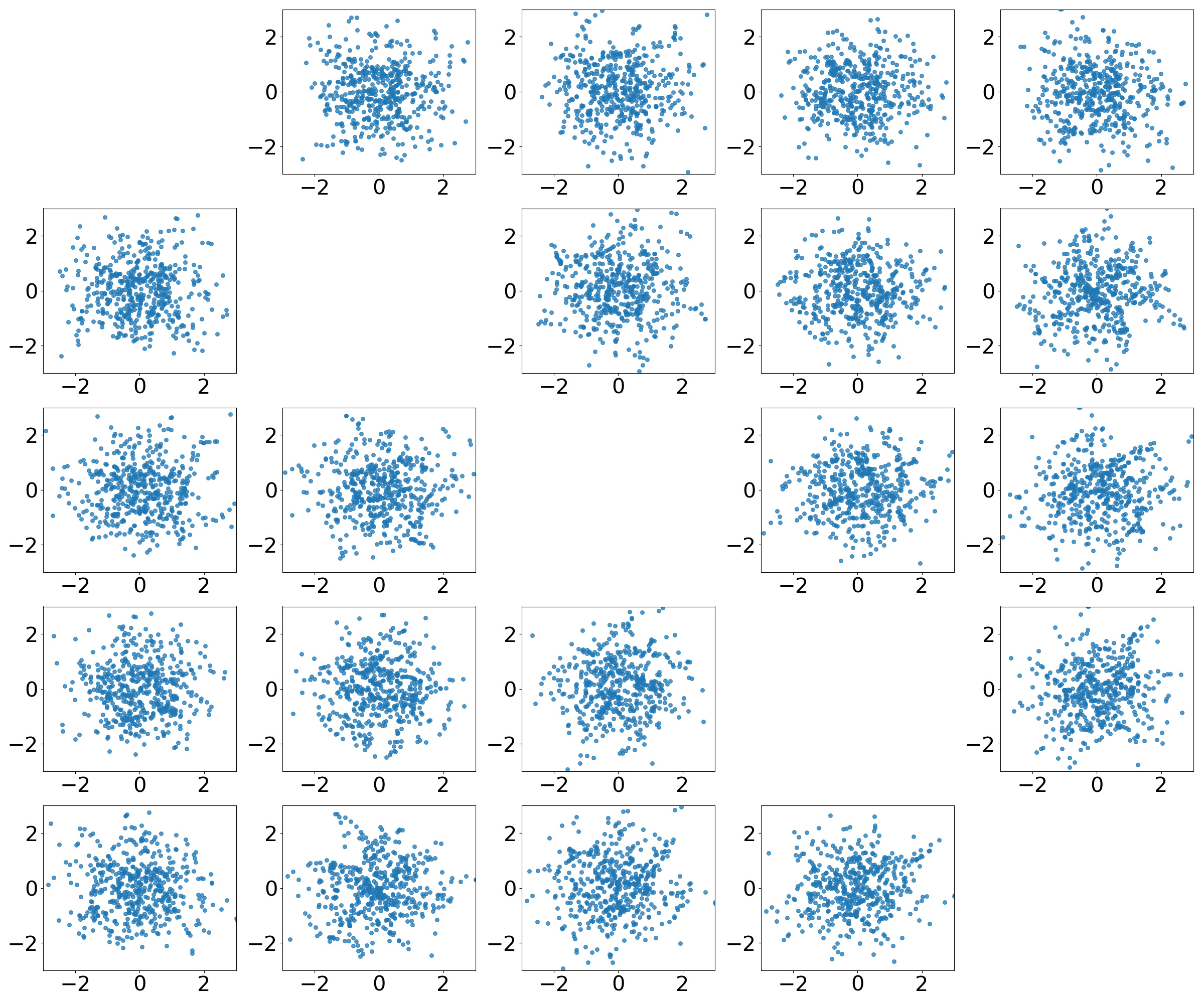}
        \caption{Swiss Roll}
        \label{fig:latent_sr}
    \end{subfigure}
    \hfill
    \begin{subfigure}{0.3\textwidth}
        \centering
        \includegraphics[width=1.0\linewidth, height=2.25in]{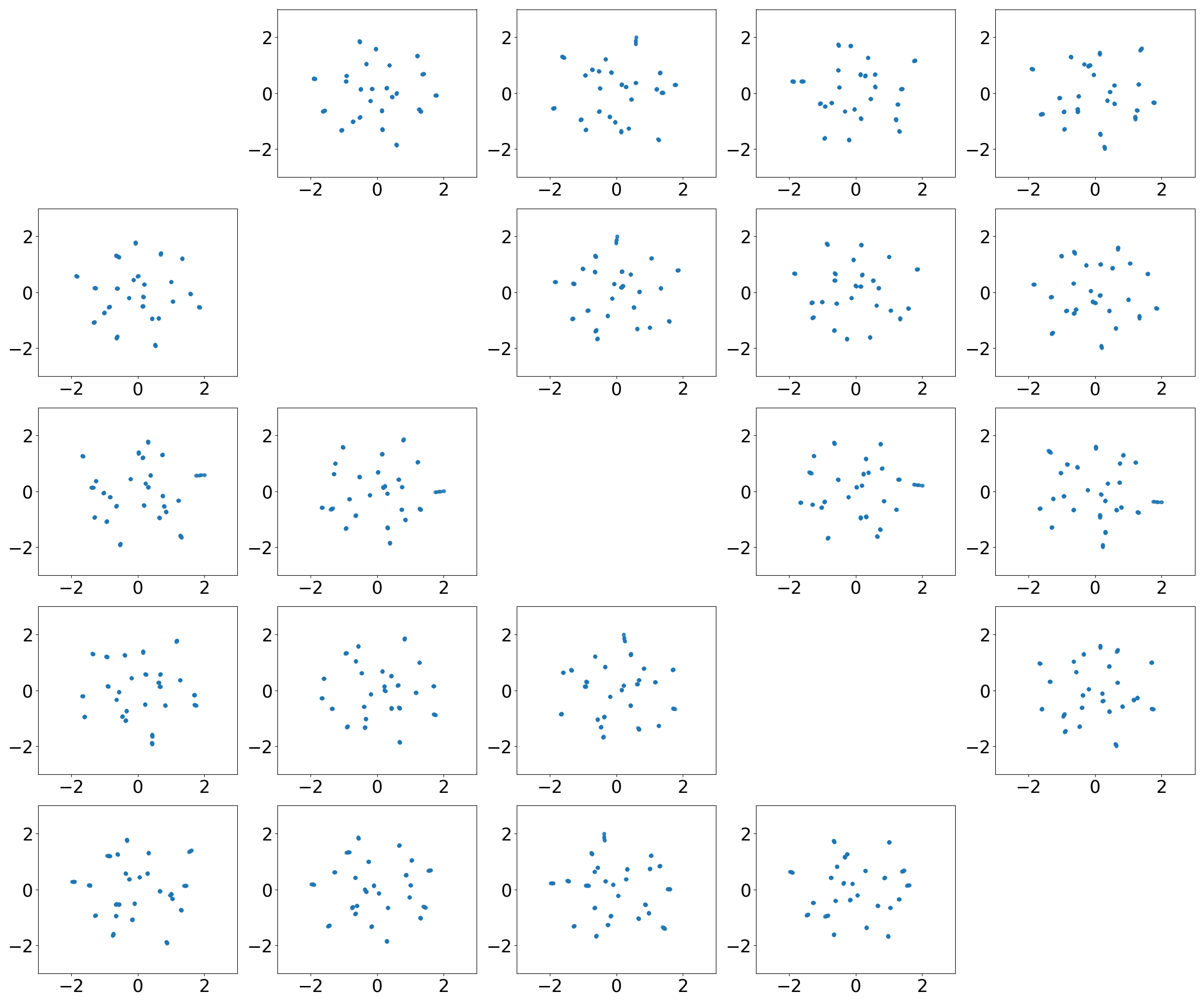}
        \caption{GRID}
        \label{fig:latent_grid}
    \end{subfigure}
    \caption{Latent spaces of mixture of Gaussians, i.e. $Q(x)_i$ against $Q(x)_j$ $i \neq j$. The joint distribution of any two dimensions of $Q(X)$ is close to a bivariate normal distribution.}
    \label{fig:latent}
\end{figure*}

\noindent{\bf Latent space.} We choose the latent distribution to be a 5-dimensional standard multivariate normal distribution $Z\sim N(0, I_5)$. During the training each batch size is chosen to be 512. After training, the distribution of $Q(X)$ is expected to be close to the distribution of $Z$. To demonstrate the latent distribution visually, we plot the $i$th compoment of $Q(X)$, $Q(X)_i$, against the $j$th compoment of $Q(X)$, $Q(X)_j$, for all $i \neq j$ in Figure \ref{fig:latent}. We can tell that the joint distribution of any two dimensions of $Q(X)$ is close to a bivariate normal distribution.

\begin{figure}[h!]
    \centering
    \includegraphics[width = 0.75\textwidth]{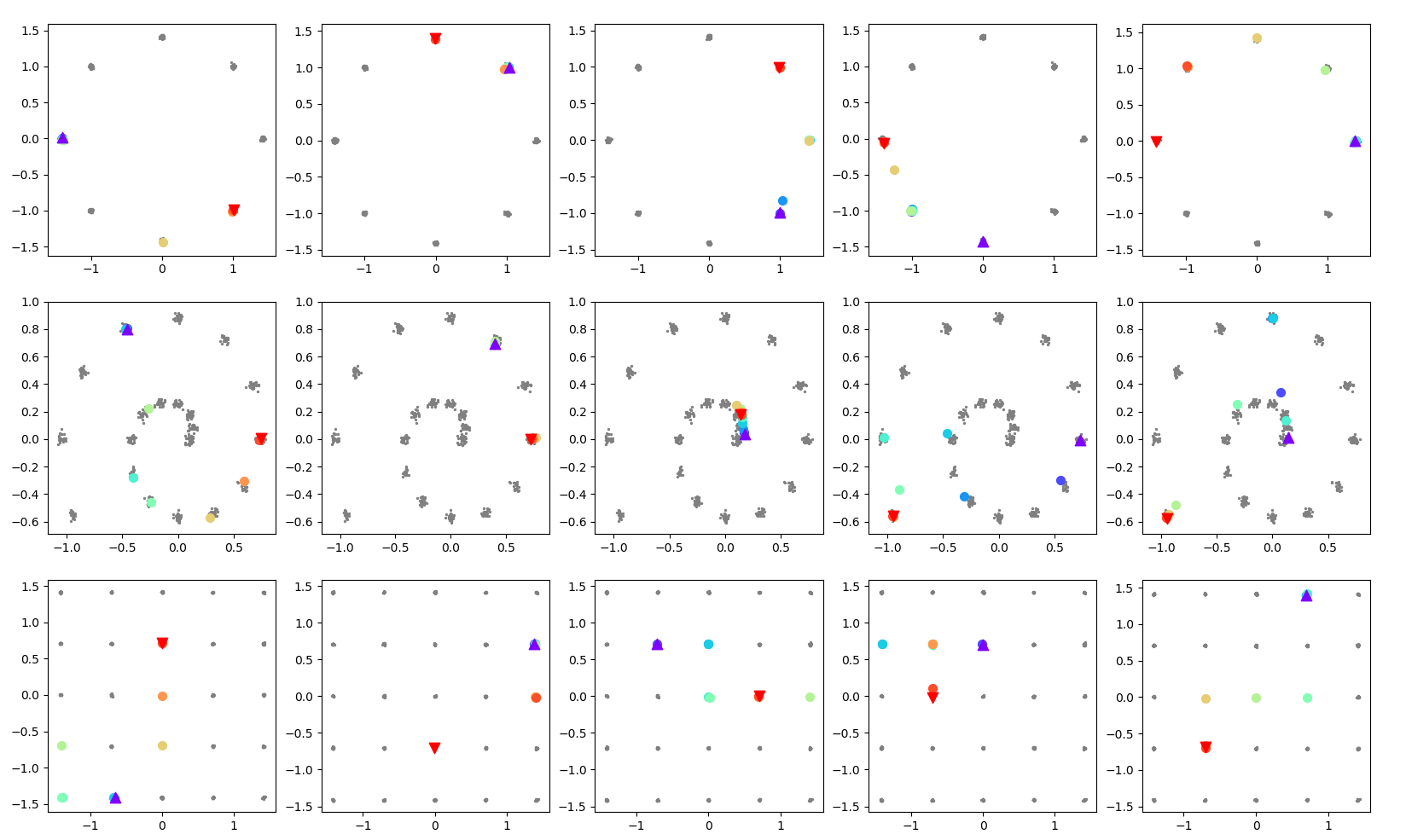}
    \caption{Interpolations: {\color{red}{$\blacktriangledown$}} and {\color{Orchid}$\blacktriangle$} indicates the first and last samples in the interpolations, other colored samples are the interpolations. The first sample is generated by $G(z_1)$ and the last sample is generated by $G(z_2)$, and other colored dots are generated by $G(\lambda z_1 + (1-\lambda) z_2)$, where $\lambda \in (0, 1)$. This figure indicates almost all interpolated samples are falling around one of the modes and successfully avoid gaps in between modes.}
    \label{fig:intp}
\end{figure}

\noindent\textbf{Mode collapse.} We investigate the mode collapse problem for the iWGAN. If we draw two random samples in the latent space $z_1, z_2\sim N({0}, {I_5})$, the interpolation, $G(\lambda z_1 + (1-\lambda) z_2)$, $0 \leq \lambda \leq 1$, should fall around the mode to represent a reasonable sample. In Figure \ref{fig:intp}, we select $\lambda \in \{0, 0.05, 0.10, \dots, 0.95, 1.0\}$, and do interpolations on two random samples. We repeat this procedure several times on 3 datasets as demonstrated in Figure \ref{fig:intp}. No matter where the interpolations start and end, the interpolations would fall around the modes other than the locations where true distribution has a low density. There may still be some samples that appears in the middle of two modes. This may be because the generator $G$ is not able to approximate a step function well.

\begin{figure}[h!]
    \centering
    \includegraphics[width=0.9\textwidth]{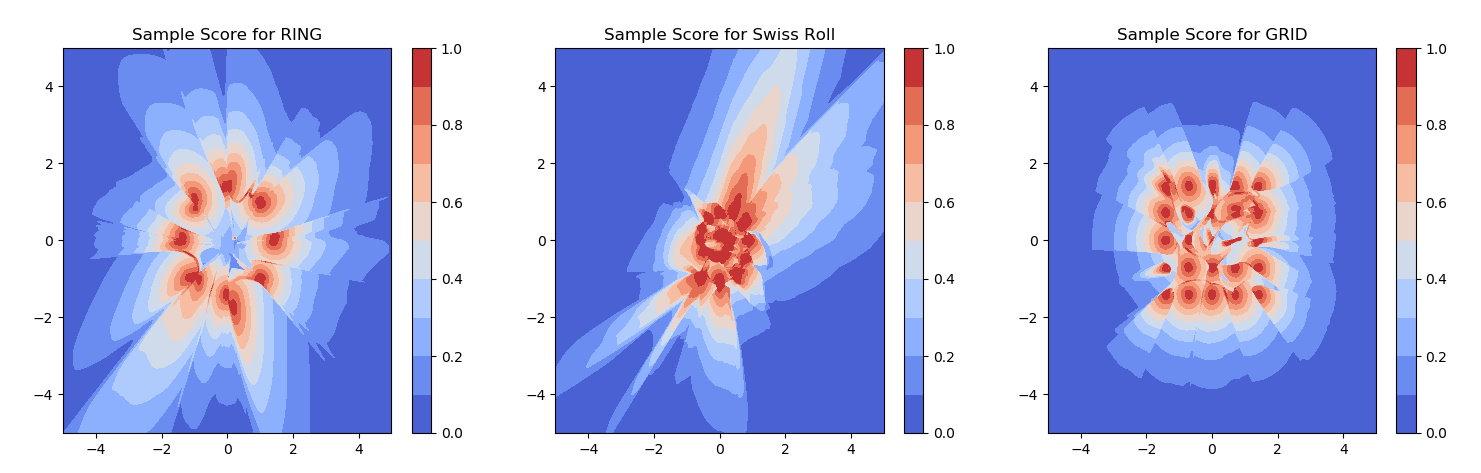}
    \caption{Quality check with the heatmap of quality scores over the space of $X$}
    \label{fig:quality}
\end{figure}

\noindent{\bf Individual sample quality check.} From the probability interpretation of the iWGAN, we naturally adopt the reconstruction error $\|X - G(Q(X))\|$, or the {\it quality score} 
\[\mbox{Quality Score} = \exp{(-\|X-G(Q(X))\|)}\] 
as the metric of the quality of any individual sample. The larger the quality score is, the better quality the sample has. Figure \ref{fig:quality} shows their quality scores for different samples. The quality scores of samples near the modes of the true distribution are close to 1, and become smaller as the sample draw is away from the modes. This indicates that the iWGAN converges and learns the distribution well, and the {\it quality score} is a reliable metric for the individual sample quality.

\subsection{CelebA}\label{sec:celeba}
We experimentally demonstrate our model’s ability on two well-known benchmark datasets, MNIST and CelebA. We present the performance of the iWGAN on CelebA in this section and the performance on  MNIST  in the Appendix. CelebA (CelebFaces Attributes Dataset) is a large-scale face attributes dataset with $202,599$ $64\times 64$ colored celebrity face images, which cover large pose variations and diverse people. This dataset is ideal for training models to generate synthetic images. The MNIST database (Modified National Institute of Standards and Technology database) is another large database of handwritten digits $0\sim 9$ that is commonly used for training various image processing systems.  The MNIST database contains $70,000$ $28\times 28$ grey images. CelebA is a more complex dataset than MNIST. The CelebA dataset is available at \url{http://mmlab.ie.cuhk.edu.hk/projects/CelebA.html} and the MNIST dataset is available at \url{http://yann.lecun.com/exdb/mnist/}. 

\begin{figure}[t]
    \centering
    \begin{subfigure}[t]{0.23\textwidth}
        \centering
        \includegraphics[height=1.6in]{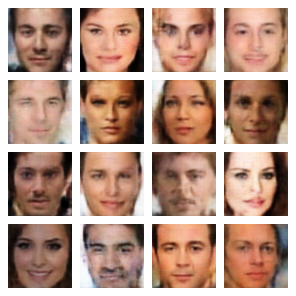}
        \caption{Generated samples}
        \label{fig:random_celeba}
    \end{subfigure}
    \hspace{0.05in}
    \begin{subfigure}[t]{0.23\textwidth}
        \centering
        \includegraphics[height=1.6in]{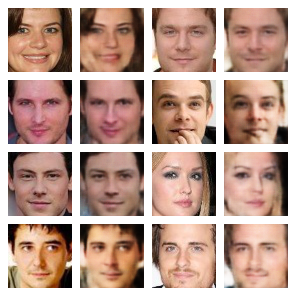}
        \caption{Reconstructions}
        \label{fig:reconst_celeba}
    \end{subfigure}
    \begin{subfigure}[t]{0.5\textwidth}
        \centering
        \includegraphics[height=1.6in]{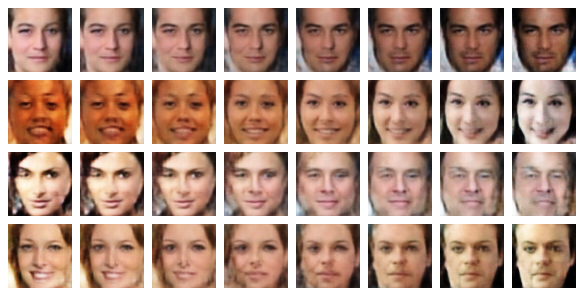}
        \caption{Interpolations}
        \label{fig:intplts_celeba}
    \end{subfigure}
    \caption{The iWGAN on CelebA. (a) The samples are generated by $G(z)$, where $z \sim N(\textbf{0}, I_{64})$; (b) The first and third columns are the original images and the second and fourth columns are the corresponding reconstructed images generated by $G(Q(x))$; (c) Four examples of interpolation between two random sample from latent space $z_1$ and $z_2$, the interpolations are generated by $G((1-\lambda) z_1+\lambda z_2)$}.\label{fig:iwgan_celeba}
\end{figure}

\begin{figure}[t]
    \centering
    \includegraphics[width=0.5\textwidth]{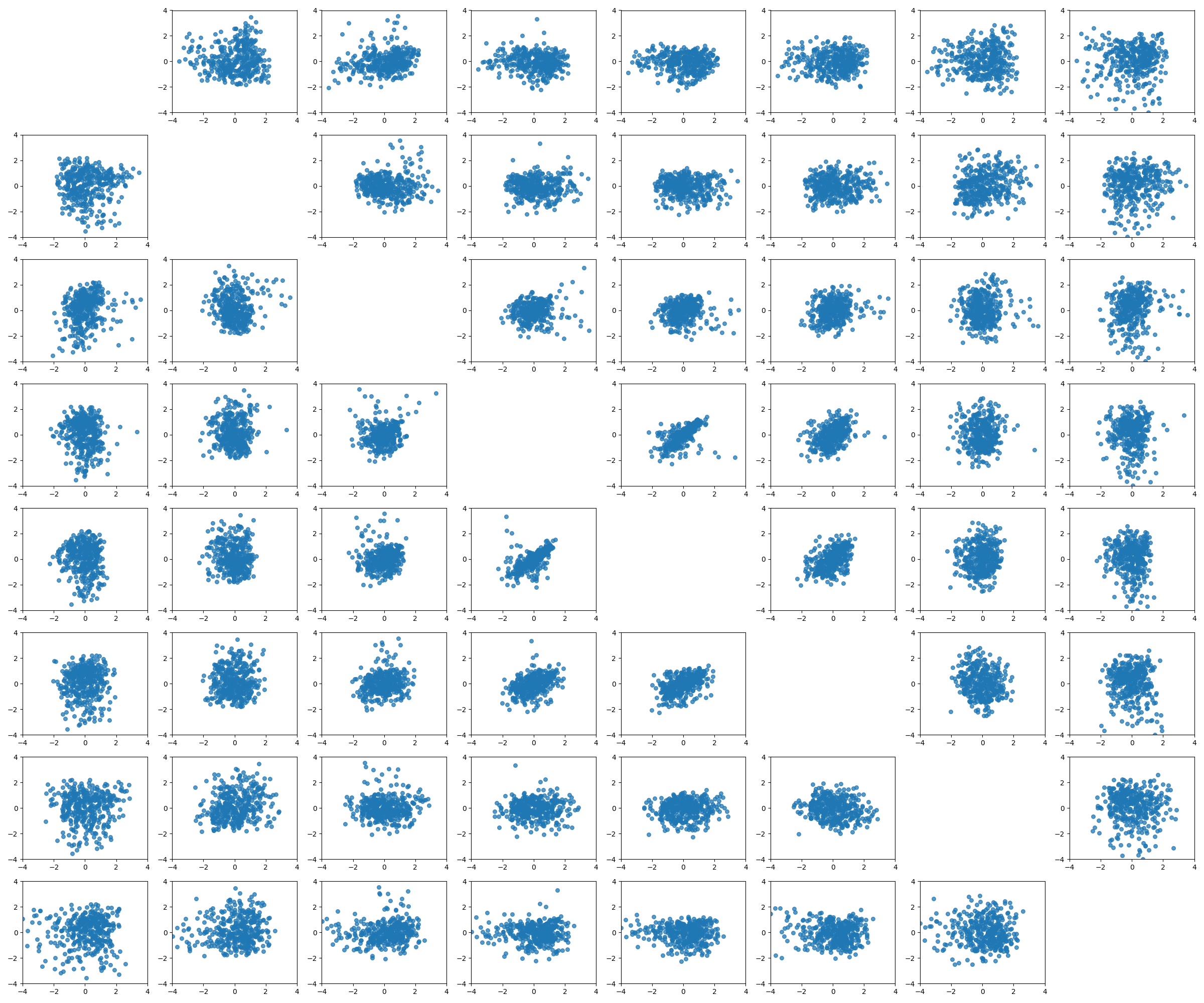}
    \caption{Latent space (first 8 dimensions) of CelebA, i.e. $Q(x)_i$ against $Q(x)_j$ $i \neq j$. The joint distribution of any two dimensions of $Q(X)$ is close to a bivariate normal distribution.}
    \label{fig:latent_celeba}
\end{figure}

\begin{figure}[!ht]
    \centering
    \includegraphics[width=0.4\textwidth]{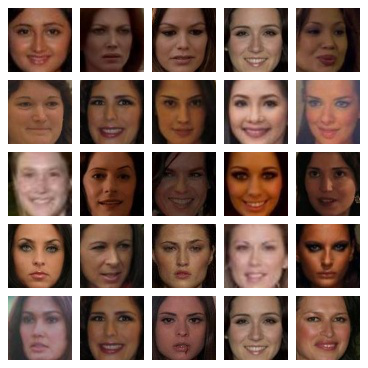}
    \hspace{0.4in}
    \includegraphics[width=0.4\textwidth]{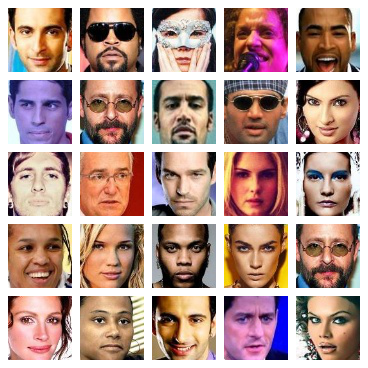}
    \caption{Images from CelebA with high (left) and low (right) quality scores by the iWGAN.}\label{fig:celeba_quality}
\end{figure}

The first result by the iWGAN  on CelebA is shown in Figure \ref{fig:iwgan_celeba}. The dimension of the latent space is chosen to be $64$. For each panel, Figure \ref{fig:iwgan_celeba} respectively shows the generated samples from $G(Z)$, the reconstructed samples from $G(Q(X))$, and the latent space interpolation between two randomly chosen images. In particular, we perform latent space interpolations between CelebA validation set examples. We sample pairs of validation set examples $x_1$ and $x_2$ and project them into $z_1$ and $z_2$ by the encoder $Q$. We then linearly interpolate between $z_1$ and $z_2$ and pass the intermediate points through the decoder to plot the input-space interpolations. In addition, Figure \ref{fig:latent_celeba} shows the first 8 dimensions of the latent space calculated by $Q(x)$ on CelebA. Figures \ref{fig:iwgan_celeba} and \ref{fig:latent_celeba} visually demonstrate that the iWGAN can simultaneously generate high quality samples, produce small reconstruction errors, and have meaningful latent codes. Figure \ref{fig:celeba_quality} also displays images with high and low quality scores selected from CelebA. The images with low quality scores are quite different with other images in the dataset and these images usually contain lighter background with masks or glasses. 

\begin{figure}[!ht]
    \centering
    \begin{subfigure}{0.3\textwidth}
        \centering 
        \includegraphics[width=\textwidth]{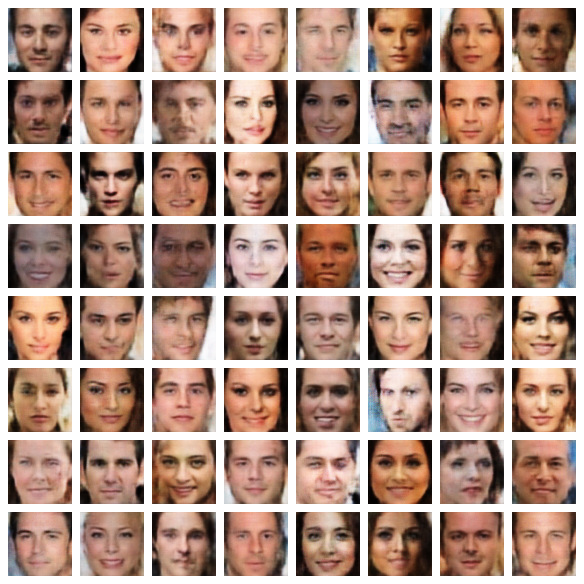}
        \caption{iWGAN}
        \label{fig:iwgan_celeba_app}
    \end{subfigure}
    \quad
    \begin{subfigure}{0.3\textwidth}
        \centering
        \includegraphics[width=\textwidth]{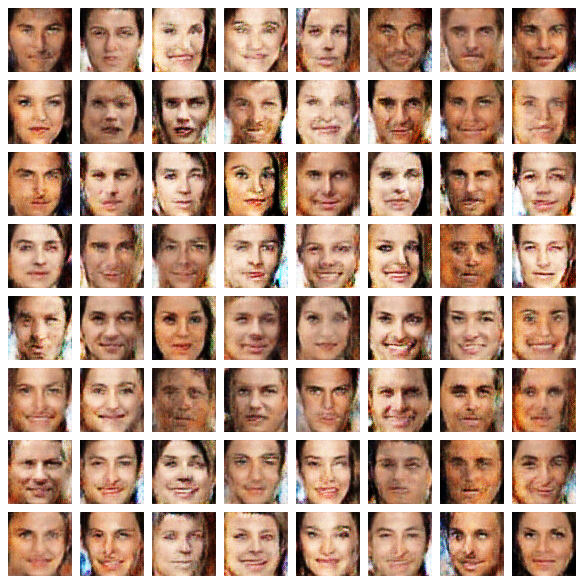}
        \caption{WGAN-GP}
        \label{fig:wgan_celeba_app}
    \end{subfigure}
    \quad
    \begin{subfigure}{0.3\textwidth}
        \centering       \includegraphics[width=\textwidth]{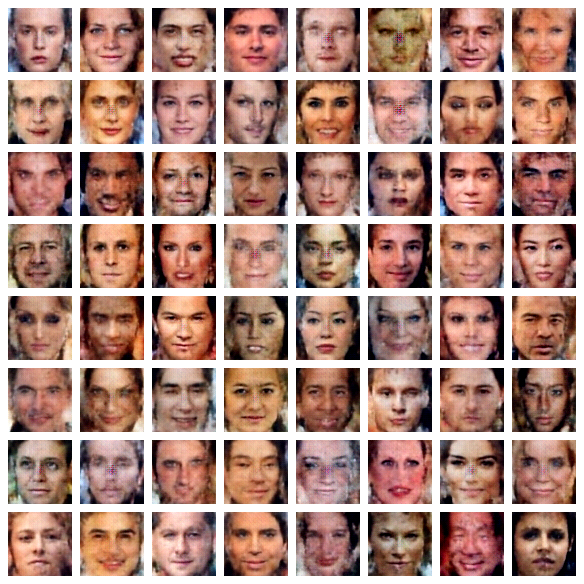}
        \caption{WAE}
        \label{fig:wae_celeba_app}
    \end{subfigure}
    \medskip
    \begin{subfigure}{0.3\textwidth}
        \centering       
        \includegraphics[width=\textwidth]{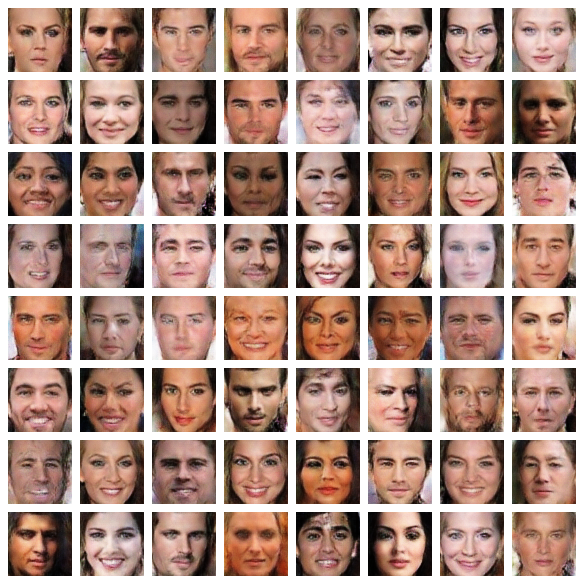}
        \caption{ALI}
        \label{fig:ali_celeba_app}
    \end{subfigure}
    \quad
    \begin{subfigure}{0.3\textwidth}
        \centering       
        \includegraphics[width=\textwidth]{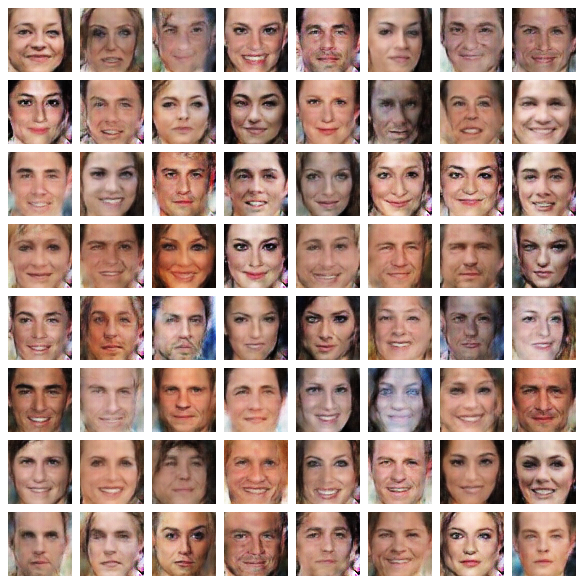}
        \caption{CycleGAN}
        \label{fig:cyclegan_celeba_app}
    \end{subfigure}
    \caption{Comparison of generated samples by different models.}
\end{figure}

\begin{figure}[!ht]
\begin{center}
\begin{subfigure}{0.75\textwidth}
    \includegraphics[width=0.95\textwidth]{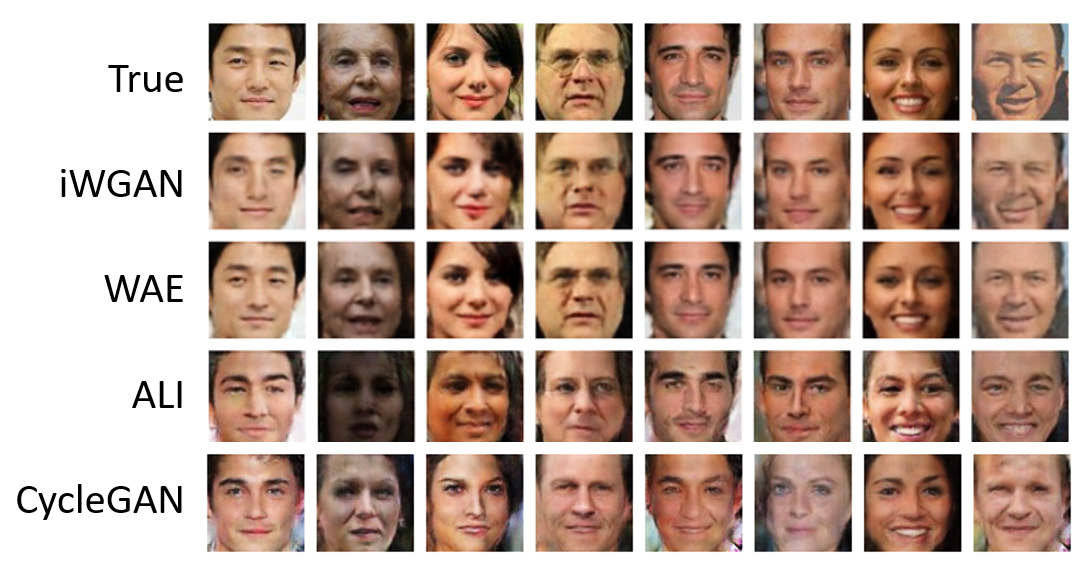}
    \caption{Reconstructions}
    \label{fig:reconstructed_celeba_app}
    \end{subfigure}
\begin{subfigure}{0.75\textwidth}
    \includegraphics[width=0.95\textwidth]{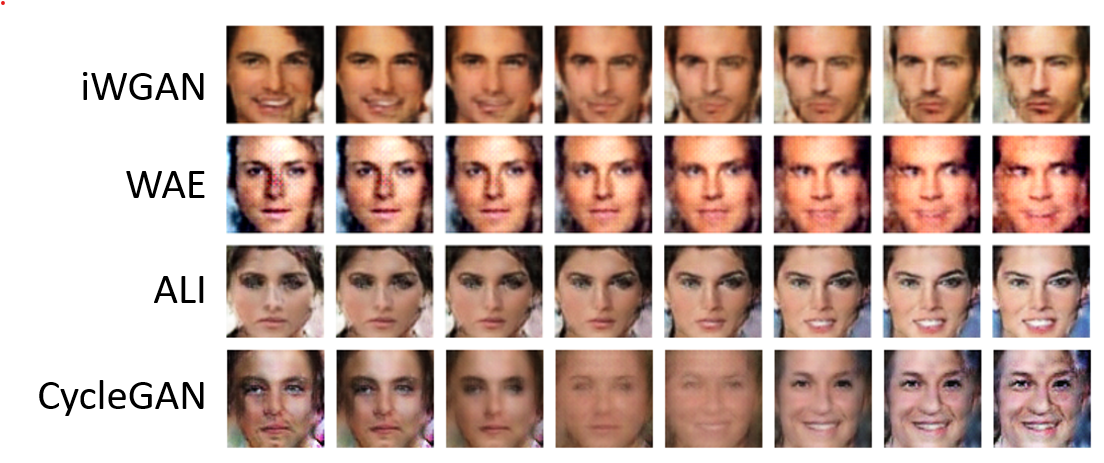}
    \caption{Interpolations}
    \label{fig:intplts_celeba_app}
    \end{subfigure}
    \end{center}
    \caption{Reconstructions and interpolations by different models.}\label{fig:iwgan_re}
\end{figure}

We compare the iWGAN, both visually and numerically, with the WGAN-GP, WAE, ALI, and CycleGAN. Figures \ref{fig:iwgan_celeba_app}---\ref{fig:cyclegan_celeba_app} display the random generated samples from the iWGAN, WGAN-GP, WAE, ALI, and CycleGAN, respectively. The generated faces by the iWGAN demonstrate higher qualities than other four methods. The top panel of Figure \ref{fig:iwgan_re} shows the comparison between real images and reconstructed images among  four methods, the iWGAN, WAE, ALI, and CycleGAN. Note that the WGAN-GP cannot provide reconstructed images since it does not produce the latent codes. The bottom panel of Figure \ref{fig:iwgan_re} shows the interpolated images by the iWGAN, WAE, ALI, and CycleGAN.

We numerically compare these five methods, the iWGAN, ALI, WAE, CycleGAN, and WGAN-GP. Four performance measures are chosen, which are inception scores (IS), Fréchet inception distances (FID), reconstruction errors (RE), and maximum mean discrepancy (MMD) between encodings and standard normal random variables. The details of these comparison metrics are given in Appendix G. Proposed by \cite{salimans2016improved}, the IS involves using a pre-trained Inception v3 model to predict the class probabilities for each generated image. Higher scores are better, corresponding to a larger KL-divergence between the two distributions. The FID is proposed by \cite{heusel2017gans} to improve the IS by actually comparing the statistics of generated samples to real samples. For the FID, the lower the better.
However, as discussed in \cite{barratt2018note}, IS is not a reliable metric for the wellness of generated samples. This is also consistent with our experiments. Although the WAE delivers the best inception scores among five methods, the WAE also has the worst FID scores. The generated samples (Figure \ref{fig:wae_celeba_app}) show that the WAE is not the best generative model compared with other four methods.
Furthermore, The reconstruction error (RE) is used to measure if the method has generated meaningful latent encodings. Smaller reconstruction errors indicate a more meaningful latent space which can be decoded into the original samples. The MMD is used to measure the difference between distribution of latent encodings and standard normal random variables. Smaller MMD indicates that the distribution of encodings is close to the standard normal distribution.

From Table \ref{tab:comparison}, in terms of generative models, the iWGAN and ALI are better models, where the WGAN-GP and CycleGAN come after, but the WAE is suffering from generating clear pictures. In terms of RE and MMD, the iWGAN and WAE are better choices, where the ALI and CycleGAN cannot always reconstruct the sample to itself (see Figure \ref{fig:reconstructed_celeba_app}). In general, Table \ref{tab:comparison} shows that the iWGAN has successfully produced both meaningful encodings and reliable generator simultaneously.

\begin{table}[t]
    \caption{Comparison of the iWGAN, ALI, WAE, CyclaGAN, WGAN-GP}
    \label{tab:comparison}
    \begin{center}
    \begin{tabular}{lrrrr}
        \hline\hline
        Methods & IS & FID & RE & MMD\\
        \hline
        True & 1.96(0.019) & 18.63 & -- & --\\
        iWGAN & 1.51(0.017) & \textbf{51.20} & \textbf{13.55(2.41)} & $\mathbf{6\times 10^{-3}}$\\
        ALI & 1.50(0.014) & \textbf{51.12} & 34.49(8.23) & $0.39$ \\
        WAE & \textbf{1.71(0.029)} & 77.53 & \textbf{9.88(1.42)} & $\mathbf{4\times 10^{-3}}$\\
        CycleGAN & {1.41(0.011)} & 61.78 & 31.90(0.84) & $0.30$\\
        WGAN-GP & \textbf{1.54(0.016)} & 61.39 & -- & -- \\
        \hline\hline
    \end{tabular}
    \end{center}
\end{table}

\section{Conclusion}

We have developed a novel iWGAN model, which fuses auto-encoders and GANs in a principled way. We have established the generalization error bound for the iWGAN. We have provided a solid probabilistic interpretation on the iWGAN using the maximum likelihood principle. Our training algorithm with an iterative primal and dual optimization has demonstrated an efficient and stable learning. We have proposed a stopping criteria for our algorithm and a metric for individual sample quality checking. The empirical results on both synthetic and benchmark datasets are state-of-the-art. 

We now mention several future directions for research on the iWGAN. First, in this paper, we assume the conditional distribution of $Z$ given $X$ is modeled by a point mass $q(z|x) = \delta(x-Q(x))$. It is interesting to extend this to a more flexible inference model. In addition, it is desirable to make the latent distribution more flexible, and consider a more general latent distribution such as the energy-based model \citep{gao20}. Second, we have ignored  approximation errors in our analysis by assuming the unknown mappings belong to the neural network spaces. It is interesting to incorporate the approximation errors to analyze the behavior of the iWGAN divergence. Third, one might be interested in applying the iWGAN into image-to-image translation, as the extension should be straightforward. A fourth direction is to develop a formal hypothesis testing procedure to test whether the samples generated from the iWAGN is the same as the data distribution. We are also working on incorporating  the iWGAN into the recent GAN modules such as the BigGAN \citep{brock2018large}, which can produce high-resolution and high-fidelity images. As its name suggests, the BigGAN focuses on scaling up the GAN models including more model parameters, larger batch sizes, and architectural changes.  Instead, the iWGAN is able to stabilize its training, and it is a promising idea to fuse these two frameworks together.

\appendix

\section*{Appendix}
\subsection*{A. Proof of Theorem 1}

According to the Nash embedding theorem \citep{nash56, gunther}, every $d$-dimensional smooth Riemannian manifold ${\cal X}$ possesses a smooth isometric embedding into $\mathbb R^p$ with $p=\max\{d(d+5)/2, d(d+3)/2+5\}$. Therefore, there exists an injective mapping  $u: {\cal X}\rightarrow \mathbb R^{p}$ which preserves the metric in the sense that the manifold metric on ${\cal X}$ is equal to the pullback of the usual Euclidean metric on $\mathbb R^p$ by $u$. The mapping $u$ is injective so that we can define the inverse mapping $u^{-1}: u(\cal X) \rightarrow {\cal X}$.

Let $\tilde X = u(X)\in  {\mathbb R}^{p}$, and write $\tilde X=(\tilde X_1, \ldots, \tilde X_p)$. Let $F_i(x)=\mathbb P(\tilde X_i\le x)$, $i=1, \ldots, p$, be the marginal cdfs. By applying the probability integral transformation to each component, the random vector
\[
\big(U_1, U_2, \ldots, U_p\big) := \big(F_1(\tilde X_1), F_2(\tilde X_2), \ldots, F_p(\tilde X_p)\big)
\]
has uniformly distributed marginals. Let $C: [0, 1]^p \rightarrow [0, 1]$ be the copula of $\tilde X$, which is defined as the joint cdf of $(U_1, \ldots, U_p)$:
\[
C(u_1, u_2, \ldots, u_p) = \mathbb P\big(U_1\le u_1, U_2\le u_2, \ldots, U_p\le u_p\big). 
\]
The copula $C$ contains all information on the dependence structure among the components of $\tilde X$, while the marginal cumulative distribution functions $F_{i}$ contain all information on the marginal distributions. Therefore, the joint cdf of $\tilde X$ is
\[
H(\tilde x_1, \tilde x_2, \ldots, \tilde x_p) = C\big(F_1(\tilde x_1), F_2(\tilde x_2), \ldots, F_p(\tilde x_p)\big).
\]
Denote the  conditional distribution of $U_k$, given $U_1, \ldots, U_{k-1}$,  by
\begin{align*}
    C_k(u_k|u_1, \ldots, u_{k-1}) & = \mathbb P\big(U_k \le u_k|U_1=u_1, \ldots, U_{k-1}=u_{k-1}\big)
\end{align*}
for $k=2, \ldots, p$. 

We will construct $Q^*$ as follows. First, we obtain $\tilde X\in \mathbb R^p$ by $\tilde X = u(X)$. Second, we transform $\tilde X$ into a random vector with uniformly distributed marginals $(U_1, \ldots, U_p)$ by the marginal cdf $F_i$. Then, define $\tilde U_1 = U_1$ and
\[
\tilde U_k = C_k\big(U_k|U_1, \ldots, U_{k-1}\big), ~~~k =2,\ldots, p.
\]
One can readily show that  $\tilde U_1, \ldots, \tilde U_p$ are independent uniform random variables. This is because
\begin{align*}
    \mathbb P(\tilde U_k \le \tilde u_k: k=1, \ldots, p) &= \int_{C_1\big(v_1\big) \le \tilde u_1}\cdots \int_{C_p\big(v_p|v_1, \ldots, v_{p-1}\big) \le \tilde u_p} dC_p\big(v_p|v_1, \ldots, v_{p-1}\big) \cdots dC_1\big(v_1\big)\\
    & = \int_{0}^{\tilde u_1} \cdots \int_0^{\tilde u_p} dz_p \cdots dz_1 = \prod_{k=1}^p \tilde u_k.
\end{align*}
In fact, this transformation is the well-known Rosenblatt transform \citep{rosenblatt52}. Finally, let $Z_i = \Phi^{-1}(\tilde U_i)$ for $i=1, \ldots, p$, where $\Phi^{-1}$ is the inverse cdf of a standard normal random variable. This completes the transformation $Q^*$ from $X$ to $Z=(Z_1, \ldots, Z_p)$. 

The above process can be inverted to obtain $G^*$. First, we transform $Z$ into independent uniform random variables by $\tilde U_i=\Phi(Z_i)$ for $i=1, \ldots, p$. Next, let $U_1 = \tilde U_1$.  Define
\[
U_k = C_k^{-1}(\tilde U_k| \tilde U_1, \ldots, \tilde U_{k-1}), ~~~ i=2, \ldots, p,
\]
where $C_k^{-1}(\cdot|u_1, \ldots, u_k)$ is the inverse of $C_k$ and can be obtained by numerical root finding. Finally, let $\tilde X_i = F_i^{-1}(U_i)$ for $i=1, \ldots, p$ and $X = u^{-1}(\tilde X)$, where $u^{-1}: u(\cal X)\rightarrow {\cal X}$ is the inverse mapping of $u$. This completes the transformation $G^*$ from $Z$ to $X$.

\subsection*{B. Proof of Theorem 2}

(a). By the iWGAN objective (3), (5) holds. Since $W_1$ is a distance between two probability measures, $W_1(P_X, P_{G(Z)}) \le \overline W_1(P_X, P_{G(Z)})$. If there exists a $Q^*\in {\cal Q}$ such that $Q^*(X)$ has the same distribution as $P_Z$, we have
\begin{align*}
    \overline W_1(P_X, P_{G(Z)}) \le W_1(P_X, P_{G(Q^*(X))}) + W_1(P_{G(Q^*(X))}, P_{G(Z)}) = W_1(P_X, P_{G(Z)}).
\end{align*}
Hence, $W_1(P_X, P_{G(Z)}) = \overline W_1(P_X, P_{G(Z)})$.\\
(b). We observe that 
\[
\sup_{f}L(\widetilde G, \widetilde Q, f) = W_1(P_X, P_{\widetilde G(\widetilde Q(X))}) + W_1(P_{\widetilde G(\widetilde Q(X))}, P_{\widetilde G(Z)}).
\]
By Theorem 1, we have $\inf_{G, Q}L(G, Q, \widetilde f) \le L(G^*, Q^*, \widetilde f)= 0$ when the encoder and the decoder have enough capacities.  Therefore, the duality gap is larger than $W_1(P_X, P_{\widetilde G(\widetilde Q(X))}) + W_1(P_{\widetilde G(\widetilde Q(X))}, P_{\widetilde G(Z)})$. It is easy to see that, if $\widetilde G$ outputs the same distribution as $X$ and $\widetilde Q$ outputs the same distribution as $Z$, both the duality gap  and $\overline W_1(P_X, P_{G(Z)})$ are zeros and $X = \widetilde G(\widetilde Q(X))$ for $X \sim P_X$.

\subsection*{C. Proof of Theorem 3}

We first consider the difference between population $W_1(P_X, P_{G(Z)})$ and empirical $\widehat{W}_1(P_X, P_{G(Z)})$ given $n$ samples $S=\{x_1,\ldots, x_n\}$. Let $f_1$ and $f_2$ be their witness function respectively. Using the dual form of 1-Wassertein distance, we have
\begin{align*}
    & W_1(P_X,P_{G(Z)}) - \widehat{W}_1(P_X,P_{G(Z)})\\
    = & \E_{X\sim P_X}\{f_1(X)\} - \E_{Z\sim P_Z}\{f_1(G(Z))\} - \frac{1}{n}\sum_{i=1}^nf_2(x_i) + \E_{Z\sim P_Z}\{f_2(G(Z))\} \\
    \le & \E_{X\sim P_X}\{f_1(X)\} - \E_{Z\sim P_Z}\{f_1(G(Z))\} - \frac{1}{n}\sum_{i=1}^nf_1(x_i) + \E_{Z\sim P_Z}\{f_1(G(Z))\}  \\
    \le & \sup_f \E_{X\sim P_X}\{f(X)\}-\frac{1}{n}\sum_{i=1}^nf(x_i)\triangleq \Phi(S).
\end{align*}

Given another sample set $S'=\{x_1, \ldots, x_i',\ldots, x_n\}$, it is clear that
\begin{align*}
    \Phi(S) -\Phi(S') 
    \le \sup_{f}\frac{|f(x_i)-f(x_i')|}{n}
    \le \frac{\|x_i-x_i'\|}{n}
    \le \frac{2B}{n},
\end{align*}
where the second inequality is obtained since $f$ is 1-Lipschitz continuous function. Applying McDiamond's Inequality, with probability at least $1-\delta/2$ for any $\delta \in (0,1)$, we have
\begin{equation}\label{eq:Phi}
    \Phi(S) \le \E\{\Phi(S)\} + B\sqrt{\frac{2}{n}\log\left(\frac{2}{\delta}\right)}.
\end{equation}
By the standard technique of symmetrization in \cite{mohri2018foundations}, we have
\begin{equation}\label{eq:ExpPhi}
    \E\{\Phi(S)\} =\E\left\{\sup_{f} \E_{X\sim P_X}\{f(X)\} - \frac{1}{n}\sum_{i=1}^n f(x_i)\right\}\le 2\mathfrak{R}_n(\mathcal{F}).
\end{equation}
It has been proved in \cite{mohri2018foundations} that with probability at least $1-\delta/2$ for any $\delta\in(0,1)$, 
\begin{equation}\label{eq:Rade}
    \mathfrak{R}_n(\mathcal{F})\le\widehat{\mathfrak{R}}_n(\mathcal{F})+B\sqrt{\frac{2}{n}\log\left(\frac{2}{\delta}\right)}.
\end{equation}
Combining Equation (\ref{eq:Phi}), Equation (\ref{eq:ExpPhi}) and Equation (\ref{eq:Rade}), we have
\[
    W_1(P_X,P_{G(Z)})\le \widehat{W}_1(P_X, P_{G(Z)})+2\widehat{\mathfrak{R}}_n(\mathcal{F})+3B\sqrt{\frac{2}{n}\log\left(\frac{2}{\delta}\right)}.
\]
By Theorem 2, we have $\widehat{W}_1(P_X,P_{G(Z)})\le \widehat{\overline{W}}_{1}(P_X, P_{G(Z)})$. Thus,
\[
    W_1(P_X,P_{G(Z)})\le \widehat{\overline{W}}_{1}(P_X, P_{G(Z)})+2\widehat{\mathfrak{R}}_n(\mathcal{F})+3B\sqrt{\frac{2}{n}\log\left(\frac{2}{\delta}\right)}.
\]

\subsection*{D. Proof of Theorem 4}

(a) It is obvious that $\ell(\theta) \le {\cal L}(\theta;\tilde\theta)$ from Equation (\ref{equ:upper}). When $q(x) = p(x|\theta)$, 
\[
    \int q(x) \log{e^{-\|x - G_\theta(Q_\theta(x))\|}\over q(x)}dx = \int p(x|\theta) \log{e^{-\|x - G_\theta(Q_\theta(x))\|}\over e^{-\|x - G_\theta(Q_\theta(x))\|-V(\theta)}}dx = V(\theta).
\]
Therefore, $\ell(\theta) = {\cal L}(\theta;\theta)$.

(b) Since $\theta^{(t)}\rightarrow\hat{\theta}$ as $t\rightarrow \infty$, we have ${\cal L}(\theta^{(t+1)}; \theta^{(t)})\rightarrow{\cal L}(\hat{\theta}; \hat{\theta}) = \ell(\hat{\theta})$. This implies $\hat{\theta}$ is the MLE.

\subsection*{E. Experimental Results on MNIST}

\subsubsection*{E.1. Latent Space}
\begin{figure}[!ht]
    \centering
    \includegraphics[width=0.5\textwidth]{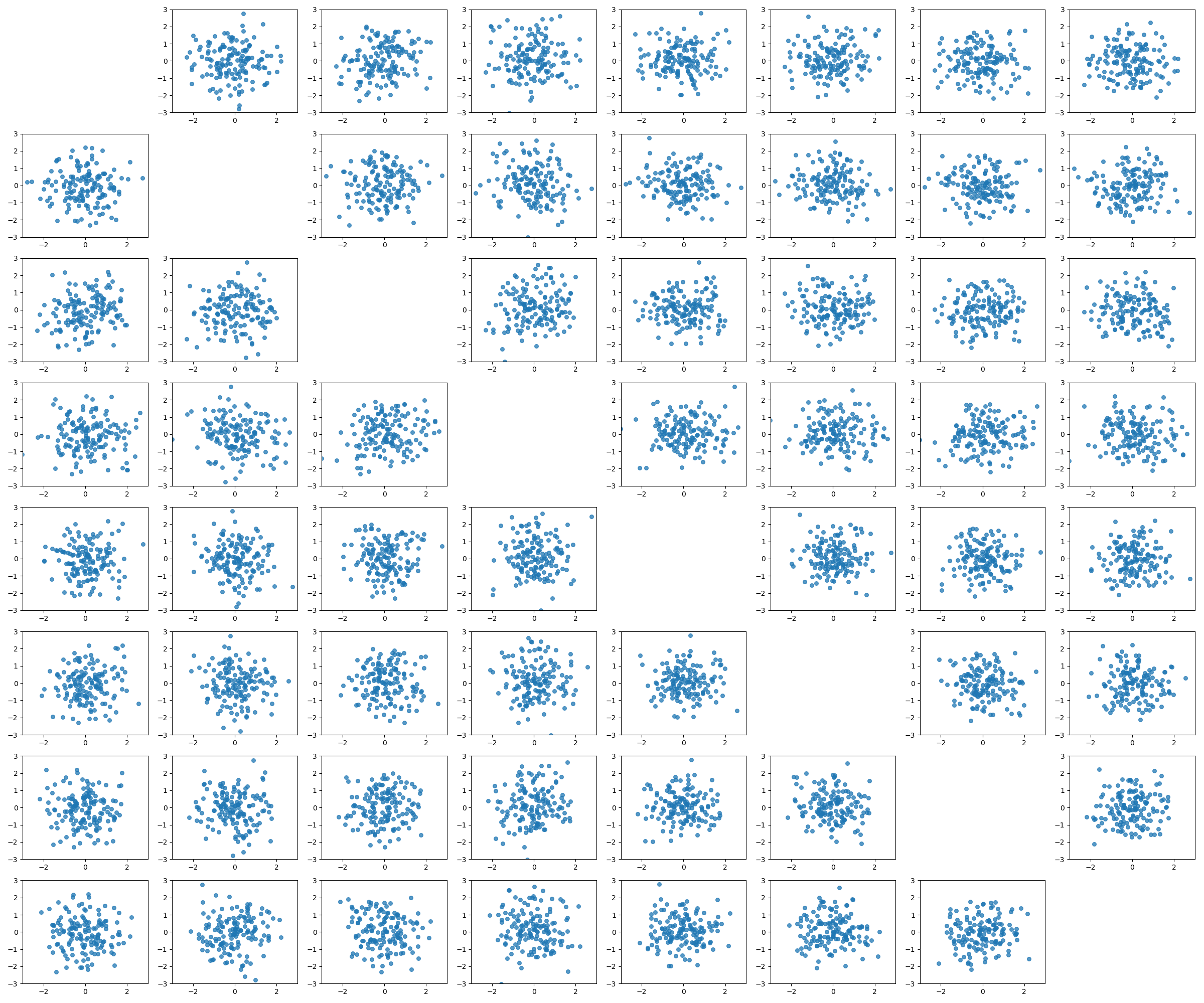}
    \caption{Latent Space of MNIST dataset}
    \label{fig:latent_mnist}
\end{figure}

Figure \ref{fig:latent_mnist} shows the latent space of MNIST, i.e. $Q(X)_i$ against $Q(X)_j$ for all $i\neq j$. 

\subsubsection*{E.2. Generated Samples}
\begin{figure}[!ht]
    \centering
    \begin{subfigure}{0.45\textwidth}
        \includegraphics[width=1.\linewidth]{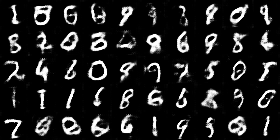}
        \caption{WGAN-GP}
        \label{fig:wgan_mnist_app}
    \end{subfigure}
    \begin{subfigure}{0.45\textwidth}
        \includegraphics[width=1.\linewidth]{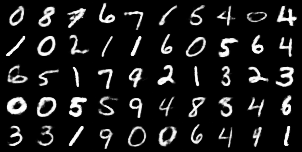}
        \caption{iWGAN}
        \label{fig:iwgan_mnist_app}
    \end{subfigure}
    \caption{Generated samples on MNIST}
    \label{fig:mnist_app}
\end{figure}

\begin{figure}[!ht]
    \centering
    \includegraphics[width=0.5\textwidth]{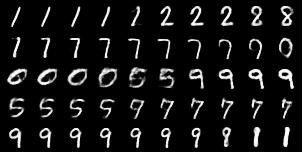}
    \caption{Interpolations by the iWGAN on MNIST}
    \label{fig:interpolation_mnist_app}
\end{figure}

Figure \ref{fig:mnist_app} shows the comparison of random generated samples between the WGAN-GP and iWGAN. Figure \ref{fig:interpolation_mnist_app} shows examples of interpolations of two random generated samples.

\subsubsection*{E.3. Reconstruction}
\begin{figure}[!ht]
    \centering
    \begin{subfigure}{0.48\textwidth}
        \includegraphics[width=1\textwidth]{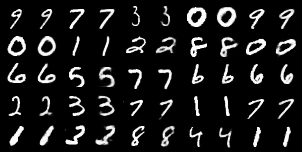}
        \caption{Reconstructions}
        \label{fig:reconstruction_mnist}
    \end{subfigure}
    \begin{subfigure}{0.45\textwidth}
        \includegraphics[width=0.95\textwidth]{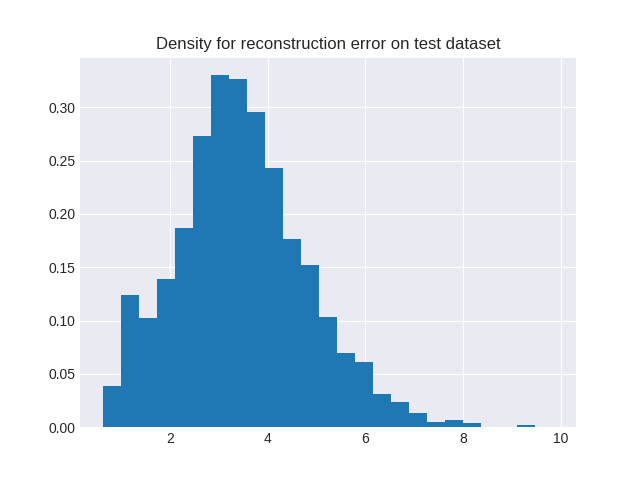}
        \vspace{-0.5cm}
        \caption{Histogram of RE}
        \label{fig:hist_mnist}
    \end{subfigure}
    \caption{Reconstructions on MNIST}
    \label{fig:reconstruct_mnist_app}
\end{figure}

\begin{figure}[!ht]
    \centering
    \begin{subfigure}{0.45\textwidth}
        \includegraphics[width=0.9\linewidth]{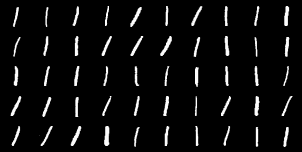}
        \caption{Samples with high quality scores}
        \label{fig:best_mnist_app}
    \end{subfigure}
    \begin{subfigure}{0.45\textwidth}
        \includegraphics[width=0.9\linewidth]{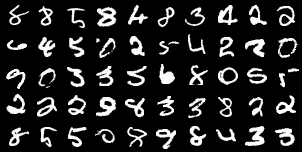}
        \caption{Samples with low quality scores}
        \label{fig:worst_mnist_app}
    \end{subfigure}
    \caption{Sample quality check by the iWGAN on the validation dataset of MNIST}
    \label{fig:mnist_quality_app}
\end{figure}

Figure \ref{fig:hist_mnist} shows, based on the samples from validation dataset, the distribution of reconstruction error. Figure \ref{fig:reconstruction_mnist} shows examples of reconstructed samples. Figure \ref{fig:mnist_quality_app} shows the best and worst samples based on quality scores from the validation dataset.

\subsection*{F. Architectures}
The codes and examples used for this paper is available at: \url{https://drive.google.com/drive/folders/1-_vIrbOYwf2BH1lOrVEcEPJUxkyV5CiB?usp=sharing}. In this section, we present the architectures used for each experiment.

\subsubsection*{Mixture of Guassians}
For Mixture Guassians, the latent space $Z \in \mathbb{R}^5$, for each batch, the sample size is 256.

Encoder architecture:
\begin{align*}
    x \in \mathbb R^2 & \rightarrow FC_{1024} \rightarrow RELU \\
    & \rightarrow FC_{512} \rightarrow RELU \\
    & \rightarrow FC_{256} \rightarrow RELU \\
    & \rightarrow FC_{128} \rightarrow RELU \rightarrow FC_{5}\\
\end{align*}

Generator architecture:
\begin{align*}
    z \in \mathbb R^5 & \rightarrow FC_{512} \rightarrow RELU \\
    & \rightarrow FC_{512} \rightarrow RELU \\
    & \rightarrow FC_{512} \rightarrow RELU \rightarrow FC_{2}\\
\end{align*}

Discriminator architecture:
\begin{align*}
    x \in \mathbb R^2 & \rightarrow FC_{512} \rightarrow RELU \\
    & \rightarrow FC_{512} \rightarrow RELU \\
    & \rightarrow FC_{512} \rightarrow RELU \rightarrow FC_{1}\\
\end{align*}

\subsubsection*{MNIST}
For MNIST, the latent space $Z \in \mathbb{R}^8$ and batch size is 250.

Encoder architecture:
\begin{align*}
    x \in \mathbb R^{28\times 28} & \rightarrow Conv_{128} \rightarrow RELU \\
    & \rightarrow Conv_{256} \rightarrow RELU \\
    & \rightarrow Conv_{512} \rightarrow RELU \rightarrow FC_{8} \\
\end{align*}

Generator architecture:
\begin{align*}
    z \in \mathbb R^{8} & \rightarrow FC_{4\times4\times512} \rightarrow RELU\\
    & \rightarrow ConvTrans_{256} \rightarrow RELU \\
    & \rightarrow ConvTrans_{128} \rightarrow RELU \rightarrow ConvTrans_{1}\\
\end{align*}

Discriminator architecture:
\begin{align*}
    x \in \mathbb R^{28\times 28} & \rightarrow Conv_{128} \rightarrow RELU \\
    & \rightarrow Conv_{256} \rightarrow RELU \\
    & \rightarrow Conv_{512} \rightarrow RELU \rightarrow FC_{1} \\
\end{align*}

\subsubsection*{CelebA}
For CelebA, the latent space $Z \in \mathbb{R}^{64}$ and batch size is 64.

Encoder architecture:
\begin{align*}
    x \in \mathbb R^{64\times64\times3} & \rightarrow Conv_{128} \rightarrow LeakyRELU \\
    & \rightarrow Conv_{256} \rightarrow InstanceNorm \rightarrow LeakyRELU \\
    & \rightarrow Conv_{512} \rightarrow InstanceNorm \rightarrow LeakyRELU \rightarrow Conv_{1} \\
\end{align*}

Generator architecture:
\begin{align*}
    z \in \mathbb R^{64} & \rightarrow FC_{4\times4\times1024} \\
    & \rightarrow ConvTrans_{512} \rightarrow BN \rightarrow RELU \\
    & \rightarrow ConvTrans_{256} \rightarrow BN \rightarrow RELU \\
    & \rightarrow ConvTrans_{128} \rightarrow BN \rightarrow RELU \rightarrow ConvTrans_{3} \\
\end{align*}

Discriminator architecture:
\begin{align*}
    x \in \mathbb R^{64\times64\times3} & \rightarrow Conv_{128} \rightarrow LeakyRELU \\
    & \rightarrow Conv_{256} \rightarrow InstanceNorm \rightarrow LeakyRELU \\
    & \rightarrow Conv_{512} \rightarrow InstanceNorm \rightarrow LeakyRELU \rightarrow Conv_{1} \\
\end{align*}

\subsection*{G. Comparison Metrics}\label{sec:metric}
Four performance measures, such as inception scores (IS), Fréchet inception distances (FID), reconstruction errors (RE), and maximum mean discrepancy (MMD) between encodings and standard normal random variables, are used to compare different models. 

Proposed by \cite{salimans2016improved}, the IS involves using a pre-trained Inception v3 model to predict the class probabilities for each generated image. These predictions are then summarized into the IS by the KL divergence as following,
\begin{equation}\label{eq:IS}
    \text{IS} = \exp\left(\mathbb{E}_{x\sim P_{G(Z)}}D_{KL}\left(p(y|x)\Vert p(y)\right)\right),
\end{equation}
where $p(y|x)$ is the predicted probabilities conditioning on the generated images, and $p(y)$ is the corresponding marginal distribution. Higher scores are better, corresponding to a larger KL-divergence between the two distributions. The FID is proposed by \cite{heusel2017gans} to improve the IS by actually comparing the statistics of generated samples to real samples. It is defined as the Fréchet distance between two multivariate Gaussians,
\begin{equation}\label{eq:FID}
    \text{FID} = \Vert\mu_r-\mu_G\Vert^2 +\text{Tr}\left(\Sigma_r+\Sigma_G-2(\Sigma_r\Sigma_G)^{1/2}\right),
\end{equation}
where $X_r\sim N(\mu_r,\Sigma_r)$ and $X_G\sim N(\mu_G, \Sigma_G)$ are the 2048-dimensional activations of the Inception-v3 pool-3 layer for real and generated samples respectively. For the FID, the lower the better.
Furthermore, the reconstruction error (RE) is defined as 
\begin{equation}\label{eq:RE}
    \mbox{RE} = \dfrac{1}{N}\sum_{i=1}^N \|\hat{X}_i - X_i\|_2,
\end{equation}
where $\hat{X}_i$ is the reconstructed sample for $X_i$. RE is used to measure if the method has generated meaningful latent encodings. Smaller reconstruction errors indicate a more meaningful latent space which can be decoded into the original samples.
The maximum mean discrepancy (MMD) is defined as
\begin{equation}\label{eq:MMD}
    \mbox{MMD} = \dfrac{1}{N(N-1)} \sum_{l\neq j} k(z_l, z_j) + \dfrac{1}{N(N-1)}\sum_{l\neq j} k(\tilde{z}_l, \tilde{z}_j) - \dfrac{2}{N^2} \sum_{l, j} k(z_l, \tilde{z}_j)
\end{equation}
where $k$ is  a positive-definite reproducing kernel, $z_i$'s are drawn from prior distribution ${P}_Z$, and $\tilde{z}_i = Q(x_i)$ are the latent encodings of real samples. MMD is used to measure the difference between distribution of latent encodings and standard normal random variables. Smaller MMD indicates that the distribution of encodings is close to the standard normal distribution.

\bibliography{ref}

\begin{thebibliography}{}

\bibitem[\protect\citeauthoryear{Arjovsky, Chintala, and Bottou}{Arjovsky
  et~al.}{2017}]{arjovsky2017wasserstein}
Arjovsky, M., S.~Chintala, and L.~Bottou (2017).
\newblock Wasserstein generative adversarial networks.
\newblock In {\em International conference on machine learning}, pp.\
  214--223. PMLR.

\bibitem[\protect\citeauthoryear{Arora, Ge, Liang, Ma, and Zhang}{Arora
  et~al.}{2017}]{arora2017generalization}
Arora, S., R.~Ge, Y.~Liang, T.~Ma, and Y.~Zhang (2017).
\newblock Generalization and equilibrium in generative adversarial nets (gans).
\newblock In {\em Proceedings of the 34th International Conference on Machine
  Learning-Volume 70}, pp.\  224--232. JMLR. org.

\bibitem[\protect\citeauthoryear{Barratt and Sharma}{Barratt and
  Sharma}{2018}]{barratt2018note}
Barratt, S. and R.~Sharma (2018).
\newblock A note on the inception score.
\newblock {\em arXiv preprint arXiv:1801.01973\/}.

\bibitem[\protect\citeauthoryear{Bartlett, Foster, and Telgarsky}{Bartlett
  et~al.}{2017}]{bartlett2017spectrally}
Bartlett, P.~L., D.~J. Foster, and M.~J. Telgarsky (2017).
\newblock Spectrally-normalized margin bounds for neural networks.
\newblock In {\em Advances in Neural Information Processing Systems}, pp.\
  6240--6249.

\bibitem[\protect\citeauthoryear{Berthelot, Schumm, and Metz}{Berthelot
  et~al.}{2017}]{berthelot2017began}
Berthelot, D., T.~Schumm, and L.~Metz (2017).
\newblock Began: Boundary equilibrium generative adversarial networks.
\newblock {\em arXiv preprint arXiv:1703.10717\/}.

\bibitem[\protect\citeauthoryear{Blei, Kucukelbir, and McAuliffe}{Blei
  et~al.}{2017}]{blei17}
Blei, D.~M., A.~Kucukelbir, and J.~D. McAuliffe (2017).
\newblock Variational inference: A review for statisticians.
\newblock {\em Journal of the American statistical Association\/}~{\em
  112\/}(518), 859--877.

\bibitem[\protect\citeauthoryear{Brock, Donahue, and Simonyan}{Brock
  et~al.}{2019}]{brock2018large}
Brock, A., J.~Donahue, and K.~Simonyan (2019).
\newblock Large scale {GAN} training for high fidelity natural image synthesis.
\newblock In {\em International Conference on Learning Representations}.

\bibitem[\protect\citeauthoryear{Carreira-Perpinan and
  Hinton}{Carreira-Perpinan and Hinton}{2005}]{carreira2005contrastive}
Carreira-Perpinan, M.~A. and G.~E. Hinton (2005).
\newblock On contrastive divergence learning.
\newblock In {\em Aistats}, Volume~10, pp.\  33--40. Citeseer.

\bibitem[\protect\citeauthoryear{Chen, Wang, and Ge}{Chen
  et~al.}{2018}]{chen2018training}
Chen, X., J.~Wang, and H.~Ge (2018).
\newblock Training generative adversarial networks via primal-dual subgradient
  methods: A lagrangian perspective on {GAN}.
\newblock In {\em International Conference on Learning Representations}.

\bibitem[\protect\citeauthoryear{Donahue, Kr{\"a}henb{\"u}hl, and
  Darrell}{Donahue et~al.}{2017}]{donahue2016adversarial}
Donahue, J., P.~Kr{\"a}henb{\"u}hl, and T.~Darrell (2017).
\newblock Adversarial feature learning.
\newblock In {\em International Conference on Learning Representations (ICLR)}.

\bibitem[\protect\citeauthoryear{Dumoulin, Belghazi, Poole, Mastropietro, Lamb,
  Arjovsky, and Courville}{Dumoulin et~al.}{2017}]{dumoulin2017adversarially}
Dumoulin, V., I.~Belghazi, B.~Poole, O.~Mastropietro, A.~Lamb, M.~Arjovsky, and
  A.~Courville (2017).
\newblock Adversarially learned inference.
\newblock In {\em International Conference on Learning Representations (ICLR)}.

\bibitem[\protect\citeauthoryear{Farnia and Tse}{Farnia and
  Tse}{2018}]{farnia2018convex}
Farnia, F. and D.~Tse (2018).
\newblock A convex duality framework for gans.
\newblock In {\em Advances in Neural Information Processing Systems}, pp.\
  5248--5258.

\bibitem[\protect\citeauthoryear{Fischer and Igel}{Fischer and
  Igel}{2010}]{fischer2010empirical}
Fischer, A. and C.~Igel (2010).
\newblock Empirical analysis of the divergence of gibbs sampling based learning
  algorithms for restricted boltzmann machines.
\newblock In {\em International Conference on Artificial Neural Networks}, pp.\
   208--217. Springer.

\bibitem[\protect\citeauthoryear{Gao, Nijkamp, Kingma, Xu, Dai, and Wu}{Gao
  et~al.}{2020}]{gao20}
Gao, R., R.~Nijkamp, D.~Kingma, Z.~Xu, A.~Dai, and Y.~Wu (2020).
\newblock Flow contrastive estimation of energy-based models.
\newblock In {\em 2020 IEEE/CVF Conference on Computer Vision and Pattern
  Recognition (CVPR)}, pp.\  7515--7525.

\bibitem[\protect\citeauthoryear{Goodfellow, Pouget-Abadie, Mirza, Xu,
  Warde-Farley, Ozair, Courville, and Bengio}{Goodfellow
  et~al.}{2014}]{goodfellow2014generative}
Goodfellow, I., J.~Pouget-Abadie, M.~Mirza, B.~Xu, D.~Warde-Farley, S.~Ozair,
  A.~Courville, and Y.~Bengio (2014).
\newblock Generative adversarial nets.
\newblock In {\em Advances in neural information processing systems}, pp.\
  2672--2680.

\bibitem[\protect\citeauthoryear{Gretton, Borgwardt, Rasch, Sch{\"o}lkopf, and
  Smola}{Gretton et~al.}{2012}]{gretton2012kernel}
Gretton, A., K.~M. Borgwardt, M.~J. Rasch, B.~Sch{\"o}lkopf, and A.~Smola
  (2012).
\newblock A kernel two-sample test.
\newblock {\em Journal of Machine Learning Research\/}~{\em 13\/}(Mar),
  723--773.

\bibitem[\protect\citeauthoryear{Grnarova, Levy, Lucchi, Perraudin, Hofmann,
  and Krause}{Grnarova et~al.}{2018}]{grnarova2018evaluating}
Grnarova, P., K.~Y. Levy, A.~Lucchi, N.~Perraudin, T.~Hofmann, and A.~Krause
  (2018).
\newblock Evaluating gans via duality.
\newblock {\em arXiv preprint arXiv:1811.05512\/}.

\bibitem[\protect\citeauthoryear{Gu and Zhu}{Gu and Zhu}{2001}]{gu01}
Gu, M.~G. and H.~Zhu (2001).
\newblock Maximum likelihood estimation for spatial models by markov chain
  monte carlo stochastic approximation.
\newblock {\em Journal of the Royal Statistical Society B\/}~{\em 63\/}(2),
  339--355.

\bibitem[\protect\citeauthoryear{Gulrajani, Ahmed, Arjovsky, Dumoulin, and
  Courville}{Gulrajani et~al.}{2017}]{gulrajani2017improved}
Gulrajani, I., F.~Ahmed, M.~Arjovsky, V.~Dumoulin, and A.~C. Courville (2017).
\newblock Improved training of wasserstein gans.
\newblock In {\em Advances in Neural Information Processing Systems}, pp.\
  5767--5777.

\bibitem[\protect\citeauthoryear{G\"unther}{G\"unther}{1991}]{gunther}
G\"unther, M. (1991).
\newblock Isometric embeddings of riemannian manifolds.
\newblock In {\em Proceedings of the International Congress of Mathematicians},
  pp.\  1137--1143.

\bibitem[\protect\citeauthoryear{Han, Nijkamp, Fang, Hill, Zhu, and Wu}{Han
  et~al.}{2019}]{han19}
Han, T., E.~Nijkamp, X.~Fang, M.~Hill, S.-C. Zhu, and Y.~N. Wu (2019).
\newblock Divergence triangle for joint training of generator model,
  energy-based model, and inferential model.
\newblock In {\em Proceedings of IEEE Conference on Computer Vision and Pattern
  Recognition (CVPR)}, pp.\  8670--8679.

\bibitem[\protect\citeauthoryear{Heusel, Ramsauer, Unterthiner, Nessler, and
  Hochreiter}{Heusel et~al.}{2017}]{heusel2017gans}
Heusel, M., H.~Ramsauer, T.~Unterthiner, B.~Nessler, and S.~Hochreiter (2017).
\newblock Gans trained by a two time-scale update rule converge to a local nash
  equilibrium.
\newblock In {\em Advances in Neural Information Processing Systems}, pp.\
  6626--6637.

\bibitem[\protect\citeauthoryear{Hinton}{Hinton}{2002}]{hinton2002training}
Hinton, G.~E. (2002).
\newblock Training products of experts by minimizing contrastive divergence.
\newblock {\em Neural computation\/}~{\em 14\/}(8), 1771--1800.

\bibitem[\protect\citeauthoryear{Hornik}{Hornik}{1991}]{hornik1991approximation}
Hornik, K. (1991).
\newblock Approximation capabilities of multilayer feedforward networks.
\newblock {\em Neural networks\/}~{\em 4\/}(2), 251--257.

\bibitem[\protect\citeauthoryear{Hu, Yang, Salakhutdinov, and Xing}{Hu
  et~al.}{2018}]{hu2017unifying}
Hu, Z., Z.~Yang, R.~Salakhutdinov, and E.~P. Xing (2018).
\newblock On unifying deep generative models.
\newblock In {\em International Conference on Learning Representations}.

\bibitem[\protect\citeauthoryear{Jiang, Chen, Chen, Liu, Wang, and Zhao}{Jiang
  et~al.}{2019}]{jiang2018computation}
Jiang, H., Z.~Chen, M.~Chen, F.~Liu, D.~Wang, and T.~Zhao (2019).
\newblock On computation and generalization of generative adversarial networks
  under spectrum control.
\newblock In {\em International Conference on Learning Representations}.

\bibitem[\protect\citeauthoryear{Kingma and Ba}{Kingma and
  Ba}{2015}]{kingma2014method}
Kingma, D.~P. and J.~Ba (2015).
\newblock Adam: A method for stochastic optimization.
\newblock In {\em 3rd International Conference for Learning Representations,
  San Diego, 2015}.

\bibitem[\protect\citeauthoryear{Kingma and Welling}{Kingma and
  Welling}{2014}]{kingma2013auto}
Kingma, D.~P. and M.~Welling (2014).
\newblock {Auto-Encoding Variational Bayes}.
\newblock In {\em 2nd International Conference on Learning Representations,
  {ICLR} 2014, Banff, AB, Canada, April 14-16, 2014, Conference Track
  Proceedings}.

\bibitem[\protect\citeauthoryear{Larsen, S{\o}nderby, Larochelle, and
  Winther}{Larsen et~al.}{2016}]{larsen2016autoencoding}
Larsen, A. B.~L., S.~K. S{\o}nderby, H.~Larochelle, and O.~Winther (2016).
\newblock Autoencoding beyond pixels using a learned similarity metric.
\newblock In {\em International Conference on Machine Learning}, pp.\
  1558--1566.

\bibitem[\protect\citeauthoryear{Li, Lu, Wang, Haupt, and Zhao}{Li
  et~al.}{2019}]{li2018tighter}
Li, X., J.~Lu, Z.~Wang, J.~Haupt, and T.~Zhao (2019).
\newblock On tighter generalization bounds for deep neural networks: {CNN}s,
  resnets, and beyond.

\bibitem[\protect\citeauthoryear{Li, Swersky, and Zemel}{Li
  et~al.}{2015}]{li2015generative}
Li, Y., K.~Swersky, and R.~Zemel (2015).
\newblock Generative moment matching networks.
\newblock In {\em International Conference on Machine Learning}, pp.\
  1718--1727. PMLR.

\bibitem[\protect\citeauthoryear{Mescheder, Nowozin, and Geiger}{Mescheder
  et~al.}{2017}]{mescheder2017adversarial}
Mescheder, L., S.~Nowozin, and A.~Geiger (2017).
\newblock Adversarial variational bayes: Unifying variational autoencoders and
  generative adversarial networks.
\newblock In {\em Proceedings of the 34th International Conference on Machine
  Learning-Volume 70}, pp.\  2391--2400. JMLR. org.

\bibitem[\protect\citeauthoryear{Mohri, Rostamizadeh, and Talwalkar}{Mohri
  et~al.}{2018}]{mohri2018foundations}
Mohri, M., A.~Rostamizadeh, and A.~Talwalkar (2018).
\newblock {\em Foundations of machine learning}.
\newblock MIT press.

\bibitem[\protect\citeauthoryear{Nash}{Nash}{1956}]{nash56}
Nash, J. (1956).
\newblock The imbedding problem for riemannian manifolds.
\newblock {\em Annals of mathematics\/}~{\em 63\/}(1), 20--63.

\bibitem[\protect\citeauthoryear{Nowozin, Cseke, and Tomioka}{Nowozin
  et~al.}{2016}]{nowozin2016f}
Nowozin, S., B.~Cseke, and R.~Tomioka (2016).
\newblock f-gan: Training generative neural samplers using variational
  divergence minimization.
\newblock In {\em Advances in neural information processing systems}, pp.\
  271--279.

\bibitem[\protect\citeauthoryear{Qiu and Wang}{Qiu and Wang}{2020}]{qiu20jasa}
Qiu, Y. and X.~Wang (2020).
\newblock Almond: Adaptive latent modeling and optimization via neural networks
  and langevin diffusion.
\newblock {\em Journal of the American Statistical Association\/}~{\em 0\/}(0),
  1--13.

\bibitem[\protect\citeauthoryear{Qiu, Zhang, and Wang}{Qiu
  et~al.}{2020}]{qiu20}
Qiu, Y., L.~Zhang, and X.~Wang (2020).
\newblock Unbiased contrastive divergence algorithm for training energy-based
  latent variable models.
\newblock In {\em International Conference on Learning Representations (ICLR)}.

\bibitem[\protect\citeauthoryear{Rosca, Lakshminarayanan, Warde-Farley, and
  Mohamed}{Rosca et~al.}{2017}]{rosca2017variational}
Rosca, M., B.~Lakshminarayanan, D.~Warde-Farley, and S.~Mohamed (2017).
\newblock Variational approaches for auto-encoding generative adversarial
  networks.
\newblock {\em arXiv preprint arXiv:1706.04987\/}.

\bibitem[\protect\citeauthoryear{Rosenblatt}{Rosenblatt}{1952}]{rosenblatt52}
Rosenblatt, M. (1952).
\newblock Remarks on a multivariate transformation.
\newblock {\em Annals of Mathematical Statistics\/}~{\em 23\/}(3), 470--472.

\bibitem[\protect\citeauthoryear{Salimans, Goodfellow, Zaremba, Cheung,
  Radford, and Chen}{Salimans et~al.}{2016}]{salimans2016improved}
Salimans, T., I.~Goodfellow, W.~Zaremba, V.~Cheung, A.~Radford, and X.~Chen
  (2016).
\newblock Improved techniques for training gans.
\newblock In {\em Advances in neural information processing systems}, pp.\
  2234--2242.

\bibitem[\protect\citeauthoryear{Schulz, M{\"u}ller, and Behnke}{Schulz
  et~al.}{2010}]{schulz2010investigating}
Schulz, H., A.~M{\"u}ller, and S.~Behnke (2010).
\newblock Investigating convergence of restricted boltzmann machine learning.
\newblock In {\em NIPS 2010 Workshop on Deep Learning and Unsupervised Feature
  Learning}.

\bibitem[\protect\citeauthoryear{Tolstikhin, Bousquet, Gelly, and
  Schoelkopf}{Tolstikhin et~al.}{2018}]{tolstikhin2018wasserstein}
Tolstikhin, I., O.~Bousquet, S.~Gelly, and B.~Schoelkopf (2018).
\newblock Wasserstein auto-encoders.
\newblock In {\em International Conference on Learning Representations}.

\bibitem[\protect\citeauthoryear{Ulyanov, Vedaldi, and Lempitsky}{Ulyanov
  et~al.}{2018}]{ulyanov2018takes}
Ulyanov, D., A.~Vedaldi, and V.~Lempitsky (2018).
\newblock It takes (only) two: Adversarial generator-encoder networks.
\newblock In {\em Thirty-Second AAAI Conference on Artificial Intelligence}.

\bibitem[\protect\citeauthoryear{Villani}{Villani}{2008}]{villani2008optimal}
Villani, C. (2008).
\newblock {\em Optimal transport: old and new}, Volume 338.
\newblock Springer Science \& Business Media.

\bibitem[\protect\citeauthoryear{Zhao, Mathieu, and LeCun}{Zhao
  et~al.}{2017}]{zhao2016energy}
Zhao, J., M.~Mathieu, and Y.~LeCun (2017).
\newblock Energy-based generative adversarial network.
\newblock In {\em International Conference on Learning Representations (ICLR)}.

\bibitem[\protect\citeauthoryear{Zhao, Song, and Ermon}{Zhao
  et~al.}{2018}]{zhao2018information}
Zhao, S., J.~Song, and S.~Ermon (2018).
\newblock The information-autoencoding family: A lagrangian perspective on
  latent variable generative modeling.

\bibitem[\protect\citeauthoryear{Zhu, Park, Isola, and Efros}{Zhu
  et~al.}{2017}]{zhu2017unpaired}
Zhu, J.~Y., T.~Park, P.~Isola, and A.~A. Efros (2017).
\newblock Unpaired image-to-image translation using cycle-consistent
  adversarial networks.
\newblock In {\em Proceedings of the IEEE international conference on computer
  vision}, pp.\  2223--2232.

\end{thebibliography}
\bibliographystyle{chicago}

\end{document}